\documentclass{article} 
\usepackage{colm2024_conference}
\newcommand{\name}[0]{prepacking}
\newcommand{\cname}[0]{Prepacking}

\usepackage{microtype}
\usepackage{hyperref}
\usepackage{url}
\usepackage{graphicx}
\usepackage{booktabs}
\usepackage{subcaption}
\usepackage{makecell}
\usepackage{multirow}
\usepackage{enumitem}
\usepackage{hyperref}
\usepackage[]{mdframed}
\usepackage{framed}
\usepackage{tabularx}
\usepackage{url}
\usepackage{algorithm}
\usepackage{algpseudocode}
\usepackage{amsbsy}
\usepackage{caption}
\usepackage{caption}
\usepackage{subcaption}
\usepackage{fontawesome5}
\usepackage{hyperref}

\usepackage{amsmath}
\definecolor{darkblue}{rgb}{0, 0, 0.5}
\hypersetup{colorlinks=true, citecolor=darkblue, linkcolor=darkblue, urlcolor=darkblue}

\title{Prepacking: A Simple Method for Fast Prefilling and Increased Throughput in Large Language Models}

\colmfinalcopy 
\author{Siyan Zhao\thanks{Equal Contribution.} , Daniel Israel\footnotemark[1] \\
Department of Computer Science\\
University of California, Los Angeles\\
\texttt{\{siyanz,disrael\}@cs.ucla.edu} \\
\And
Guy Van den Broeck, Aditya Grover \\
Department of Computer Science\\
University of California, Los Angeles\\
\texttt{\{guyvdb,adityag\}@cs.ucla.edu}
}

\begin{document}

\maketitle

\begin{abstract}

During inference for transformer-based large language models (LLM), prefilling is the computation of the key-value (KV) cache for input tokens in the prompt prior to autoregressive generation. For longer input prompt lengths, prefilling will incur a significant overhead on decoding time. In this work, we highlight the following pitfall of prefilling: for batches containing high-varying prompt lengths, significant computation is wasted by the standard practice of padding sequences to the maximum length. As LLMs increasingly support longer context lengths, potentially up to 10 million tokens, variations in prompt lengths within a batch become more pronounced. To address this, we propose \textit{\name{}}, a simple yet effective method to optimize prefilling computation. To avoid redundant computation on pad tokens, \name{} combines prompts of varying lengths into a sequence and packs multiple sequences into a compact batch using a bin-packing algorithm. It then modifies the attention mask and positional encoding to compute multiple prefilled KV-caches for multiple prompts within a single sequence. On standard curated dataset containing prompts with varying lengths, we obtain a significant speed and memory efficiency improvements as compared to the default padding-based prefilling computation within Huggingface across a range of base model configurations and inference serving scenarios.
\end{abstract}

\begin{center}
\href{https://github.com/siyan-zhao/prepacking}{\textcolor{black}{\faGithubSquare}} 
\cname{} on GitHub
\end{center}

\section{Introduction}

Transformer-based large language models (LLMs) have emerged as a powerful general purpose tool to service natural language queries~\citep{bai2022training, touvron2023llama, achiam2023gpt}. 
As language models continue to grow in scale and their usage proliferates across various domains~\citep{eloundou2023gpts}, the capability to generate tokens with optimal speed and efficiency becomes increasingly paramount.

The challenges of optimizing LLMs are unique compared to traditional software. LLMs are useful due to their generality, which means they can receive very diverse prompts, from short questions to long summarizing tasks. Due to the quadratic runtime of a Transformer, longer prompts require much more computation than short prompts. When long and short prompt queries are requested at the same time, the challenge of LLM inference is to route the queries in a manner that more computational resources are allocated where needed. In the current LLM paradigm, this poses a dilemma that worsens with increasing model scale due to longer, more compute-demanding queries. As an example, recent efforts are aimed at expanding the context window of LLMs to accommodate up to one million tokens and beyond~\citep{reid2024gemini}. The increasing diversity and complexity of queries demand a more efficient approach to computational resource allocation than ever before.

The conventional approach to LLM inference with varied size inputs is inefficient, and it is exemplified by the Huggingface Transformers library ~\citep{wolf-etal-2020-transformers}. The Huggingface library has seen widespread adoption in the NLP community. Despite its wide use, Huggingface handles prompts of varying lengths by padding all prompts to match the length of the longest sequence and processing the batch through a Transformer model in its entirety. This results in substantial memory utilization and computational inefficiency. While LLMs are compute-bound during prefilling, they are also memory-bound during generation~\citep{kwon2023efficient},
so it is crucial to optimize memory and GPU utilization to enable efficient inference and scalability.
\begin{figure}[t]
\begin{center}
    \includegraphics[width=1\textwidth, trim={0 7cm 0 5cm}, clip]{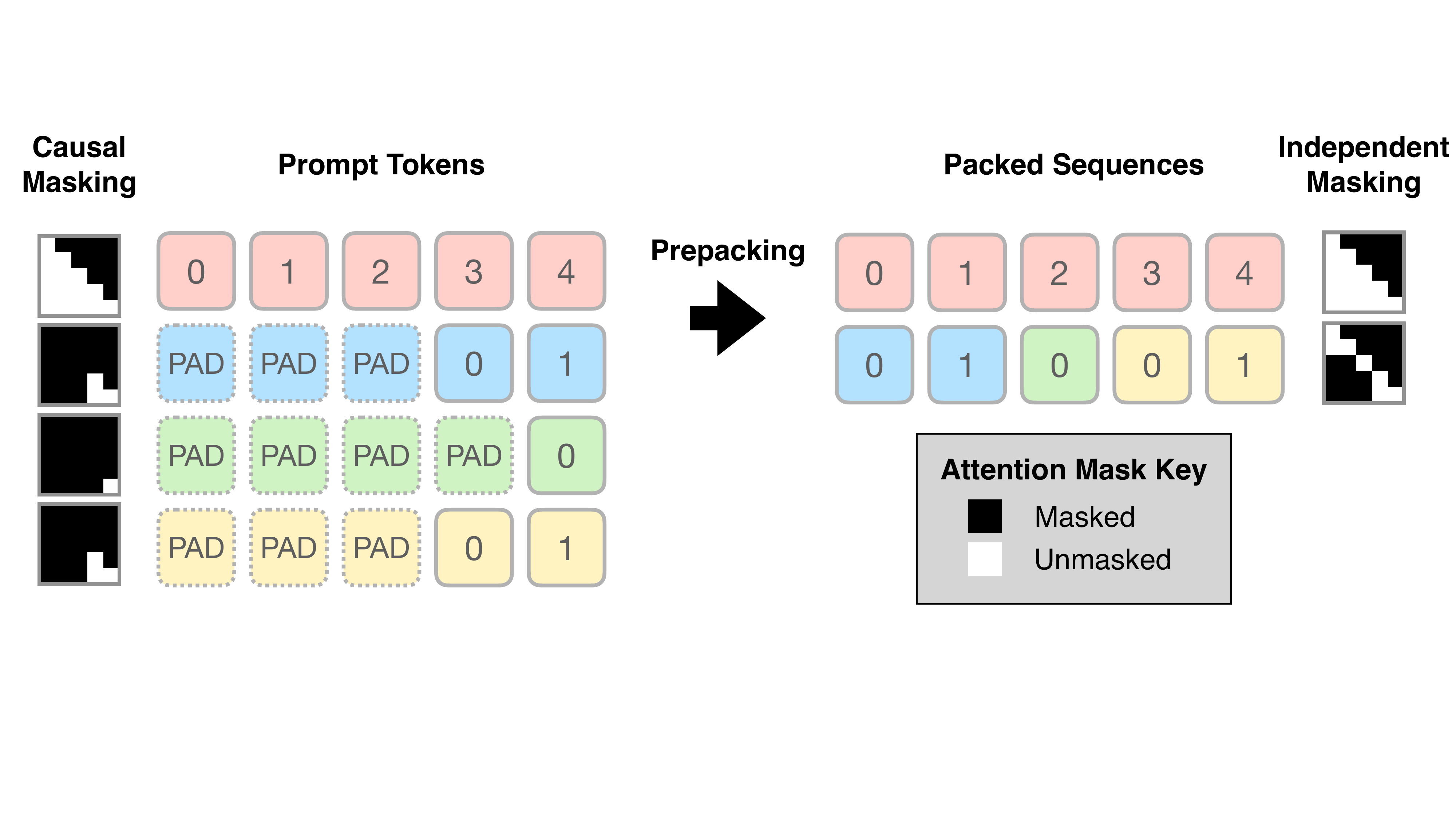}
    \caption{\textbf{Left:} The standard full batching approach (e.g., used in HuggingFace) pads shorter prompts to maximum prompt length in the batch. Each prompt has its own causal attention mask. \textbf{Right:} Prepacking combines multiple prompts into a single sequence using a bin-packing algorithm, and applies \textit{independent masking} and \textit{restart positional encodings} (numbers inside token boxes) to avoid prompts attending to other prompts. Both strategies are equivalent at decoding time, but \name{} is more compute efficient during prefilling.}
    \label{prepacking_diagram}
    \end{center}
\end{figure}

In this work, we mitigate wasteful computation with an alternative pre-processing step called \textit{\name{}}. Prepacking is specifically aimed at improving the speed and memory usage of LLM prefilling, which is the initial computation that populates the Key-Value cache (KV cache) preceding generation. Prepacking is conceptually simple; rather than padding every sequence to the same length, we pack multiple prompts together in place of padding tokens using an off-the-shelf bin-packing algorithm. This is made possible by custom attention masking and positional encoding
that enable the computation of a batch within a single sequence. The positional encoding restarts its index for each prompt in the sequence and the mask prevents prompts from attending to previous prompts in the packed sequence (Figure \ref{prepacking_diagram}). A forward pass on the pre-packed batch will populate a KV cache, which we can unpack to get the cache for the original prompts for next token generations. 

We empirically demonstrate that \name{} leads to a \textbf{speedup of up to 6x} in prefilling and time-to-first-token (TTFT) compared to the full batching method used in Huggingface tested on NVIDIA A6000 GPUs. To evaluate \name{}’s runtime performance under conditions representative of real-world user traffic, we tested it across six diverse language datasets, encompassing tasks such as question answering, summarization, instruction following, language modeling, and human preference modeling, with language models ranging from 1B to 13B parameters. \cname{} achieves greater speedup when the sequences within a batch exhibit more diverse length variations and when the batch size is large. Additionally, we demonstrate that \name{} is a simple method for increasing LLM throughput, especially in memory-constrained settings. Specifically, \name{} significantly reduces memory consumption by allowing up to \textbf{16x larger batch size} during prefilling. Beyond prefilling, we also demonstrate in premilinary experiments that \name{} can bring significant speedup and memory saving during generation.

\section{Preliminaries}
\subsection{Transformer Architecture} The decoder-only Transformer~\citep{vaswani2017attention,radford2019language} is ubiquitous in its use as the deep learning architecture for autoregressive LLMs. The core component of the Transformer is self-attention. Self-attention operates on input sequences $X\in \mathbb{R}^{n\times d}$ and is parameterized with matrices $W^Q, W^K, W^V \in \mathbb{R}^{d \times h}$.
We can write self-attention as follows
\[\text{SA}(X) = \mbox{softmax}(A) X W^V.\]
where  $A=\frac{(X W^Q)(X W^K)^\top}{\sqrt{d}}$ is an $n \times n$ attention matrix.
Thus, a Transformer forward pass will have an $\mathcal{O}(n^2)$ runtime where $n$ is the length of the input. 
To preserve autoregressive dependencies, an $n\times n$ mask $M$ is applied to $A$ such that ``past" tokens cannot attend to ``future" tokens. Finally, while attention itself is permutation-equivariant, the inputs $X$ typically incorporate positional information through the use of positional embeddings.

\subsection{Language Model Inference} Autoregressive sampling requires a forward pass through the Transformer for each new token generated. To avoid wastefully recomputing the attention matrix each forward pass, caching is standard practice at inference time. Sampling the $(n+1)$-th token autoregressively requires computing the attention matrix for $n$ previous tokens. When we generate the $(n+2)$-th token, instead of computing an $(n+1)\times (n+1)$ attention matrix, we can cache the keys and values over the first $n$ tokens to avoid redundant computation, and so on for $(n+j)$. This technique is known as KV caching~\citep{pope2023efficiently}. 

Prefilling is the population of the KV cache on the initial forward pass. In a typical text generation inference framework, a model will be run on a batch of $k$ prompts that, when tokenized, have lengths $l_1, ..., l_k$. Because a Transformer takes tensor input, the batch will be padded to the maximum length $m = \max_i l_i$. For the sake of simplicity, assume no parallelization over a batch. Although GPUs can parallelize computation over batches, we will argue in future sections that batch parallelization in practice has empirical limitations. Thus, under these assumptions the forward pass for prefilling will run in time $\mathcal{O}(km^2)$.




\subsection{Performance Metrics} Key metrics for evaluating LLM serving~\citep{miao2023towards} include latency measures such as Time-to-First-Token (TTFT), the time required for prefilling the KV cache and generating the first token, and Time-per-Output-Token (TPOT), the average time to generate each subsequent token. Together, these determine the total generation time. Throughput measures the number of requests processed per unit time. In this work, we focus on optimizing the prefilling stage by evaluating prefilling time and TTFT metrics. This is particularly important for assessing the overall responsiveness of any deployed system.
\section{\cname{}}
Although padding input prompts to the maximum length allows tensorized batch computation, the drawback is that significant computation is wasted on pad tokens. We propose a simple solution: insert more short prompts where padding was previously located. Because this method ``packs" prompts together to speed up prefilling, we refer to this method as \name{}. In formal terms, we have a set of $k$ prompts $p_1,\cdots,p_k$ of lengths $l_1,\cdots,l_k$, and our goal is to create a tensorized batch $B = (p_1', ..., p_r')$, where $p_1', ..., p_r'$ are sequences that contain the original prompts such that $r \leq k$. The full algorithm is shown in Algorithm \ref{alg:prepack}.



\begin{algorithm}
\caption{The \textit{\cname{}} Algorithm for Efficient Pre-Filling}\label{alg:prepack}
\begin{algorithmic}[1] 
\Procedure{Prepacking}{Prompts $p_1,  \cdots, p_k$, Transformer-based Language Model $\pi$} 
\State Prompt Lengths $l_1, \cdots, l_k \gets len(p_1, \cdots, p_k)$
\State Maximum Length $ m \gets \max_i l_i$
\State Packed sequences $p'_1, ..., p'_r$, bins $[idx]_{1:r} = \textsc{BINPACK}(l_1, \cdots, l_k, m)$ 
\Comment{$idx_i$ stores the start indices of the prompt(s) present in the packed sequence $p'_i$}
\State Batch $B \gets \textsc{TENSORIZE}(p'_1, ..., p'_r)$  
\State Independent Masks $[M']_{1:r} \gets \textsc{INDEPENDENT-MASK}([idx]_{1:r})$ 
\State Restart Positional Encodings $[R]_{1:r} \gets \textsc{RESTART-PE}([idx]_{1:r})$
\State Caches $KVs \gets \textsc{UNPACK}(\pi(B, [M']_{1:r}, [R]_{1:r}))$ \Comment{$\pi$ will return a KV Cache, which we unpack to obtain prompt-specific caches}
\State \textbf{return} $KVs$
\EndProcedure
\end{algorithmic}
\label{alg:prepack}
\end{algorithm}

\subsection{Bin Packing}
The problem of packing prompts together can be cast as a bin packing problem, where a bin can contain tokens from several different sequences. The goal of \name{} is to efficiently concatenate prompts together such that original prompts with lengths $l_1, ..., l_k$ are placed into the smallest possible $r$ bins, each of a fixed sized. It is guaranteed that $r \leq k$. We shall select $m$, where $m$ is the maximum prompt length as previously defined, to be the fixed size of the bins. For sequences that do not completely reach size $m$ after bin-packing, they will be padded to reach $m$.
Note that we choose the smallest possible constant for our bin size because the bin size will incur quadratic running time. In general, bin packing is an NP-hard problem \citep{garey1979computers}, but many heuristic approaches exist obtain approximate solutions \citep{buljubavsic2016consistent}. We use a First-Fit Decreasing bin packing heuristic as implemented by \cite{githubGitHubBenmaierbinpacking}.



\subsection{Prompt-wise Independent Masking and Restart Positional Encoding}
\cname{} will concatenate multiple smaller prompts under a single bin. Simply using the KV-cache of this packed sequence will be incorrect, because every prompt within the bin will attend causally to previous prompts. As a remedy, we create a custom attention mask to prevent items from attending to each other. 
We refer to this masking strategy as \textit{independent masking}. We describe our masking strategy below and illustrate it in Figure \ref{prepacking_diagram}.  

Formally, consider a causal (lower triangular) attention mask $M$ , where entry $M_{i,j}=1$ signifies that token $t_i$ can attend to $t_j$ and $i \geq j$. An independent mask $M'$ is a mask such that for all indices $a, b$ that mark the start and end of a prompt, $M'_{a:b,a:b} = L_{b-a}$, where $L_n$ is an $n \times n$ lower triangular matrix. All other entries will be 0.  Creating the attention mask and extracting the resultant KV-cache requires a certain amount of bookkeeping for tracking lengths of sequences and indices, but these operations contribute an insignificant (linear) overhead compared to the Transformer forward pass.

Lastly, we need to modify the positional encodings for the packed sequences. In general, the Transformer architecture is permutation equivariant ~\citep{naseer2021intriguing}, so the purpose of positional encodings (PE) is to give the model information about the position of a token in a sequence. Thus, in a prepacked sequence, we must edit the PEs for the tokens such that it is the same as it was in the unpacked prompts. This leads to positions that ``restart" in the packed sequence at the beginning of any new prompt, hence the name \textit{restart positional encoding}. With packed batches, independent masks, and restart PEs, we can compute and prefill the KV cache for each prompt and use it for autoregressive generation using any decoding algorithm.

\subsection{Runtime Analysis}
\label{runtime}
With Algorithm~\ref{alg:prepack}, we are guaranteed to compute the \textit{exact} KV caches as a padded, full-batching method.
Next, we analyze the gains during the prefilling stage using our approach.
Let the sum of prompt lengths over the batch be denoted by $L = \sum_i l_i$. 
In the best case scenario, our bin packing algorithm is able to pack every prompt into bins with no additional padding. Then we can express the number of bins as $r = L / m $.
We can now find the runtime of prefilling a batch with \name{} and compare it to the naive method.
\begin{align}
    \mathcal{O}(rm^2) = \mathcal{O}((L / m) m^2) = \mathcal{O}(Lm) = \mathcal{O}( km(L/k)) \leq \mathcal{O}(km^2)
\end{align}
The final inequality holds because the average length must be less than or equal to the maximum length. Also note that the \name{} algorithm itself runs in $\mathcal{O}(k \log k)$ time which is insignificant toward the overall runtime. Thus, we find that \name{} will outperform the naive padding approach in the best case scenario. In the worst case scenario, we cannot reduce the number of bins from the original batch size and $r=k$ will lead to the same runtime.  We shall show in our experiments that datasets tend to have enough length variation such that $r < k$ is a comfortable assumption in practice, and the differences between the naive method and \name{} can be stark. 

\noindent 
\begin{minipage}[b]{0.35\linewidth}
    \centering
    \includegraphics[width=\linewidth]{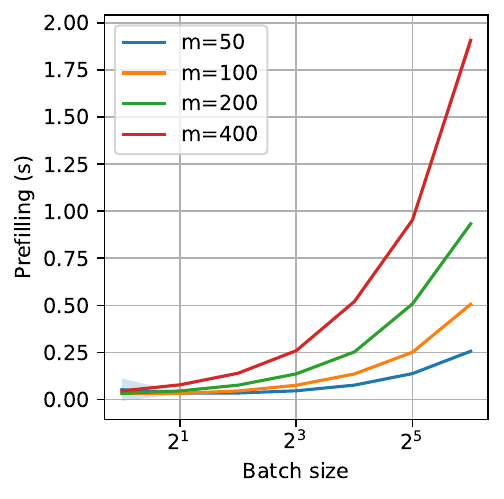}
    \captionof{figure}{Prefilling latency scaling with batch size $k$ highlights GPU parallelization limits. Results averaged over 100 runs.
    }
    \label{fig:batch_size_non_perfect_parallel}
\end{minipage}%
\hfill
\begin{minipage}[b]{0.59\linewidth}
    \centering
    \includegraphics[width=0.48\linewidth]{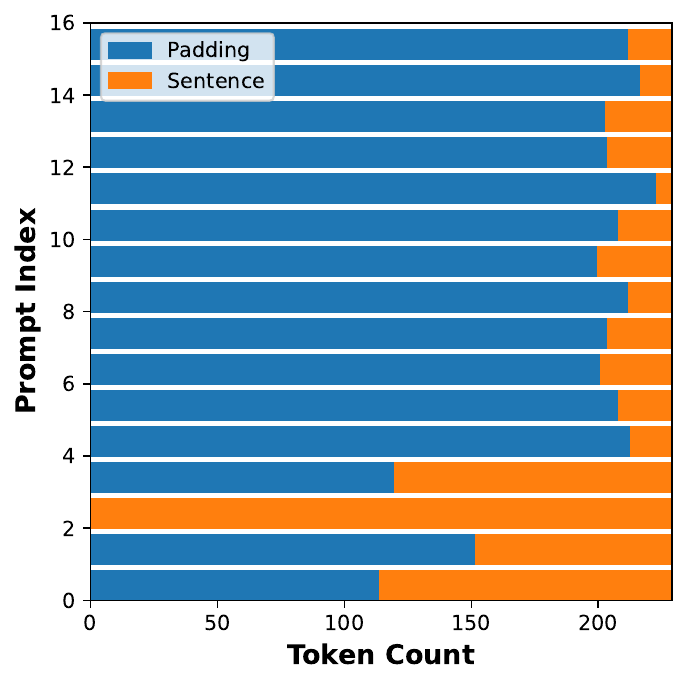}
    \hfill 
    \includegraphics[width=0.48\linewidth]{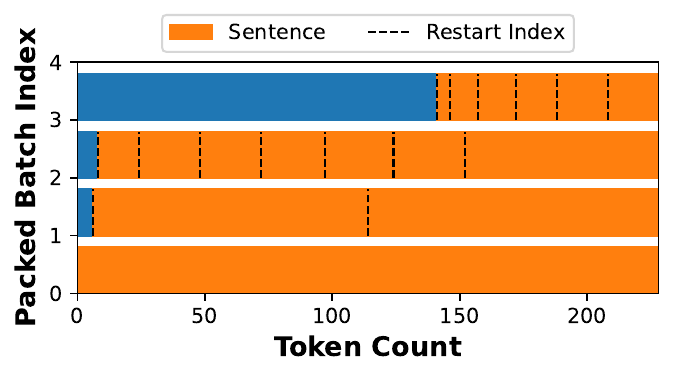}
    \captionof{figure}{An actual example of one batch sampled from the MMLU dataset shows how a batch of 16 prompts is packed into a compact packed batch. Restart indices denote the point at where independent mask and position encoding are reset to preserve the semantics of each individual prompt.}
    \label{fig:actual_pack}
\end{minipage}

\paragraph{Limitations of GPU Batch Parallelization}
Note that the above analysis assumes no parallelization over a batch. With perfect batch parallelization, \name{} will have better memory performance but no time improvement. We show empirically that GPUs cannot parallelize over batches without limitation. To show this, we sample a tensor of dimension $(k, m)$, that is batch size $k$ and prompt length $m$. In Figure \ref{fig:batch_size_non_perfect_parallel}, we demonstrate that for a fixed $m$, increasing $k$ results in a higher latency. As the batch size grows, constraints such as memory bandwidth and synchronization overhead become more pronounced~\citep{yuan2024llm}. \cname{} exploits this by reducing batch size for a fixed sequence length $m$. Figure~\ref{fig:actual_pack} illustrates an actual packing done by \name{} which greatly reduces paddings.
\section{Experiments}

We empirically show the significant throughput improvements and GPU memory savings achieved by \name{} across real-world datasets with diverse length distributions. Our comprehensive evaluation spans language models of varying architectures and scales, ranging from 1.3B to 13B parameters. With constraints on our academic budget, all experiments are conducted on a single NVIDIA 48GB A6000 GPU connected to a Colfax CX41060s-EK9 4U Rackmount Server with AMD EPYC (Genoa) 9124 processors.

\subsection{Datasets and Models}
To profile \name{}'s runtime performance under conditions representative of real-world user traffic, we evaluate on a diverse suite of datasets spanning question answering, summarization, instruction following, language modeling, and human preference modeling. Specifically, we use the MMLU~\citep{hendrycks2021ethics}, SamSum~\citep{gliwa-etal-2019-samsum}, Alpaca~\citep{alpaca}, Wikitext~\citep{merity2016pointer}, and Anthropic HH RLHF~\citep{bai2022training} datasets. While not actually evaluating task performance, we leverage the variety of formats and prompt length distributions present in these datasets to simulate the diverse input queries a LLM may encounter from user requests in production environments. Due to computational constraints, we subsample 1000 prompts from each dataset, and the lengths statistics are presented in Table~\ref{tab:datasets_length}. 
We profile a range of language models to comprehensively assess runtime impacts of scale and architecture choices: the 1.3B Sharded LLAMA~\citep{xia2023sheared}, 7B LLAMA 2~\citep{touvron2023llama} and Mistral~\citep{jiang2023mistral}, and 13B LLAMA 2~\citep{touvron2023llama} spanning 1.3B to 13B parameters with varying configurations shown in Appendix Table~\ref{tab:model_architecture}. We profile them with 4 bit or 8 bit quantization due to computational constraints. Since \name{} aims to reduce wasted computation and memory on padding within batches, for fair evaluation, we do not manually construct batches. Instead, we use actual datasets to randomly sample batches and obtain aggregate metrics with respect to diverse prompt lengths. This also reflects a more realistic setting in which the flow of queries cannot be controlled.

\subsection{Baselines}

\begin{itemize}[leftmargin=*]

    \item \textbf{\textit{Full Batching}:} As implemented by Huggingface, this method first determines the maximum prompt length across the batch and appends special padding tokens to shorter prompts until they match the maximum length. It then generates corresponding attention masks to ensure that the language model disregards the padded tokens during computation. Huggingface's inference framework~\citep{wolf-etal-2020-transformers} employs this approach for handling prompts of variable lengths, serving as the basis for this baseline's profiling.
    
    \item \textbf{\textit{Length-Ordered Batching}:} This baseline assumes access to the full set of user requests, serving as an oracle baseline that can first sort the inputs according to their lengths and sample batches in order to minimize the padding required when using the Full Batching. This method reduces computational overhead on paddings. However, it is not practical in real-world scenarios where user requests arrive in an unpredictable order, and the entire set of requests is not available upfront. In contrast, \name{} does not rely on this assumption, making it more suitable for handling dynamic and continuous streams of input prompts.

\end{itemize}

\begin{figure}[h]
\begin{center}
    \includegraphics[width=1\textwidth]{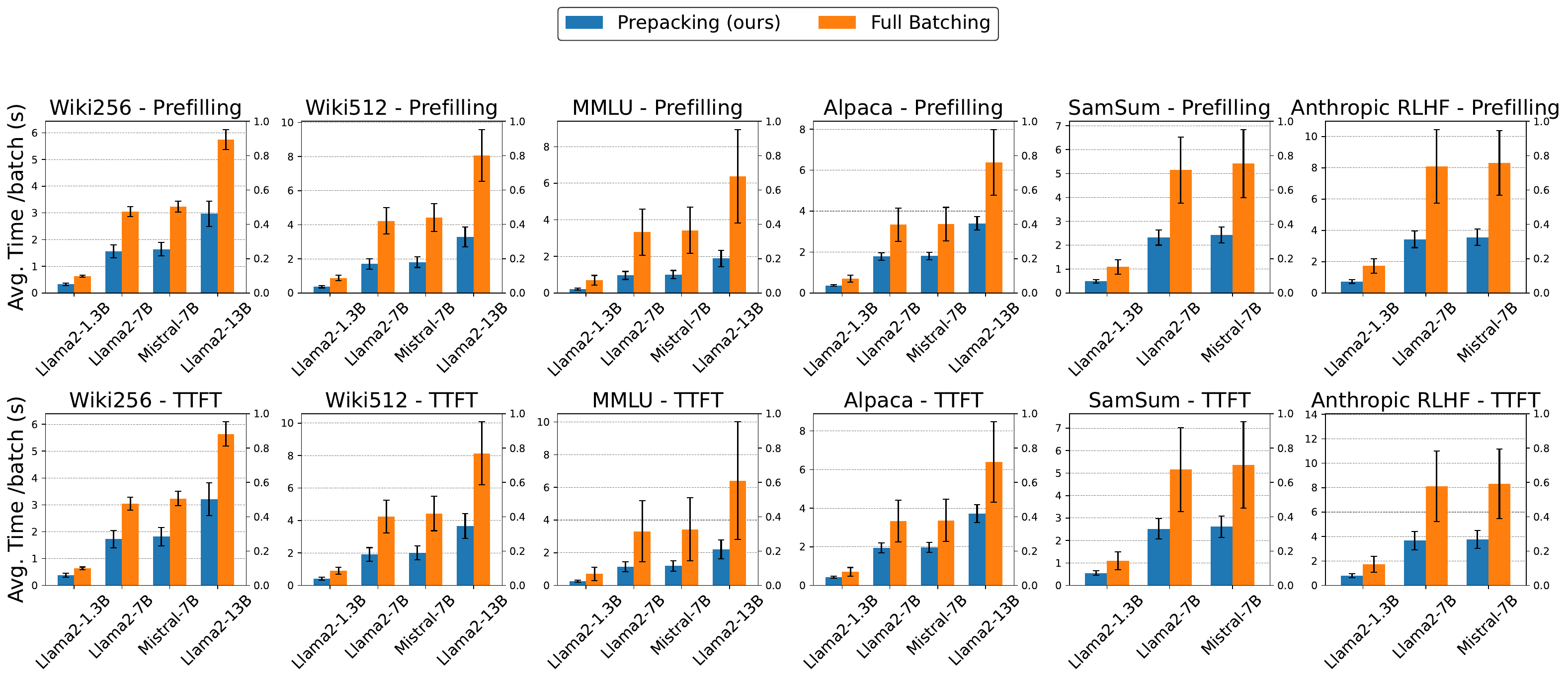}
    \caption{Average inference time per batch for various language models using \name{} and Full Batching, with a batch size of 16. The comparison is conducted across multiple datasets with two metrics, Prefilling Time and TTFT (Time-to-First-Token). Error bars represent the standard deviation of inference times across batches and seed runs. The results show that \name{} consistently leads to reduced inference times compared to Full Batching and exhibits reduced variability, as evidenced by smaller standard deviation errors, indicating more reliable and predictable inference times when adopting \name{}. }
    \label{alldataset_models_results}
    \end{center}
\end{figure}

\subsection{Prefilling Time and TTFT}
\label{sec:time}
We compare the prefilling time and Time-to-First-Token (TTFT) between \name{} and Full Batching across datasets and models in Figure~\ref{alldataset_models_results}. TTFT measures the total time required for prefilling the KV cache and generating the first token. For our method, TTFT additionally includes an overhead which is the unpacking phase, where we unpack the prompts to their original order for generation. This unpacking phase has a linear time complexity in the number of prompts, which is dominated by the quadratic computational complexity of prefilling. \cname{} consistently outperforms Full Batching with less prefilling time and TTFT, enhancing speed ranging from 1.6x to 3.5x. Moreover, \cname{} has lower inference time standard deviations, attributed to reduced padding overhead, enabling more reliable and predictable performance suitable for applications demanding consistent LLM serving.

\vspace{-2mm}
\subsection{GPU Memory Saving and Utilization}
\label{sec:memory}
\begin{figure}[h]
\begin{center}
    \includegraphics[width=0.90\textwidth]{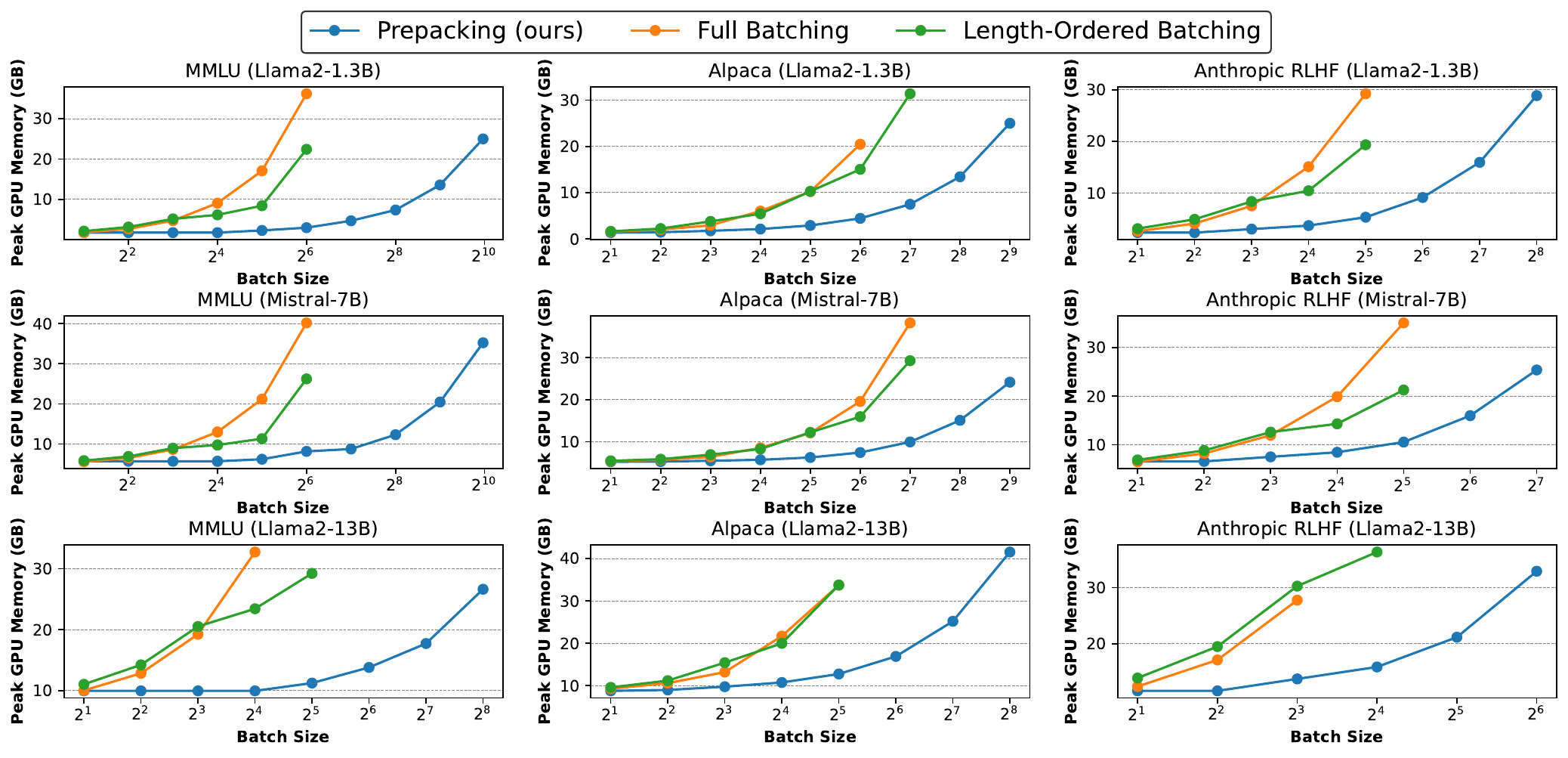}
    \caption{ Peak GPU memory usage comparison across models and datasets on a single GPU. Absent data points indicate out-of-memory errors. \cname{} achieves lower peak GPU memory usage and allows for up to \textbf{16x larger batch sizes} in prefilling computations than Full Batching and Length-Ordered Batching.} 
    \label{peakgpu}
    \end{center}
\end{figure}

We evaluate \cname{}'s GPU memory efficiency, stemming from reduced computation on padded tokens, against other baselines in Figure~\ref{peakgpu}. \cname{} consistently exhibits lower peak memory consumption, which directly translates to the ability to process larger batch sizes without encountering out-of-memory errors. For instance, with the Llama2-1.3B model on the MMLU dataset, \name{} can accommodate batch sizes up to 16x larger during prefilling compared to Full Batching before encountering OOM. This has significant implications for deploying models in resource-constrained environments, where maximizing hardware utilization is crucial. Consequently, as shown in Appendix Figure~\ref{meangpuutil}, \cname{} also exhibits lower GPU utilization when operating with the same batch size as the baselines, owing to its reduced computational overhead.

\vspace{-2mm}
\subsection{Enhanced Speedup with Increasing Batch Sizes}
\label{sec:bs_ablation}
\begin{figure}[h]
\begin{center}
    \includegraphics[width=0.85\textwidth]{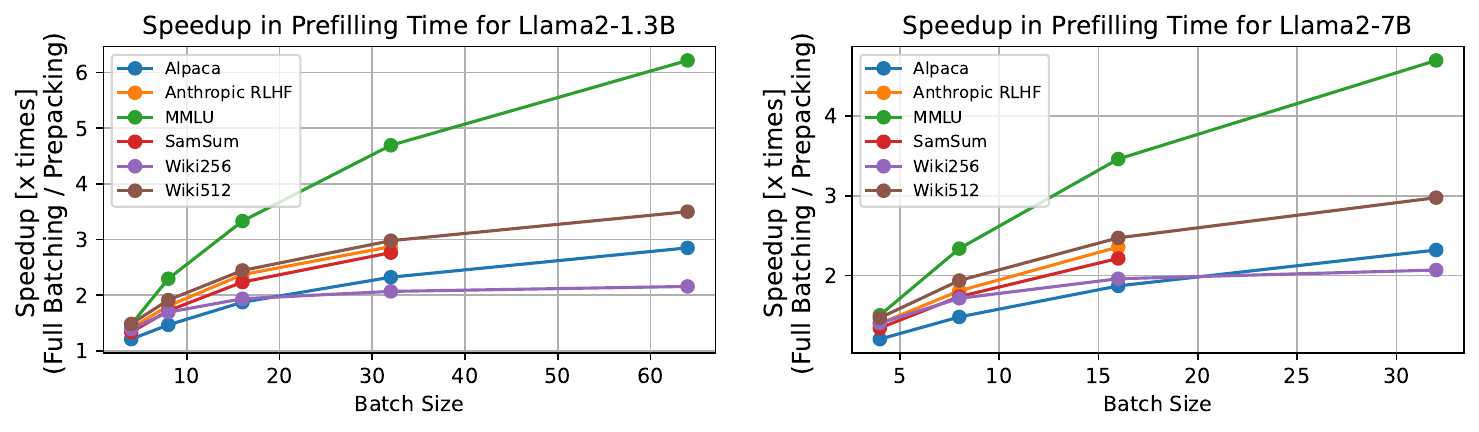}
\caption{Speed up across various batch sizes for Llama2-1.3B and Llama2-7B. Speed up is calculated as the ratio of the prefilling time with full batching to that of \name{}. Missing data points for Anthropic RLHF and SamSum are due to out-of-memory issues as these two datasets have larger mean prompt length.}
    \label{ablation_bs}
    \end{center}
\end{figure}

In realistic situations, the distribution of batch sizes encountered during language model inference can fluctuate due to non-uniform user requests arrival patterns. To evaluate our method's effectiveness in handling this variability, we conducted experiments across a range of batch sizes for the Llama2-7B and Llama2-1.3B models. The results shown in Figure~\ref{ablation_bs} demonstrate substantial speedup gains achieved by our approach over Full Batching. Larger batch sizes exhibit greater performance improvements with our method, up to \textbf{4.5x} and \textbf{6x} speedup for the 7B and 1.3B Llama2 models, respectively. This trend stems from the increased likelihood of diverse prompt lengths within larger batches, which leads to more padding overhead for Full Batching. In contrast, our method efficiently handles variable-length prompts via bin-packing, mitigating this overhead.

\subsection{Dataset \cname{} vs Length-Ordered Batching}
\label{sec:datasetprepack}
\begin{figure}[h]
\begin{center}
    \includegraphics[width=\textwidth]{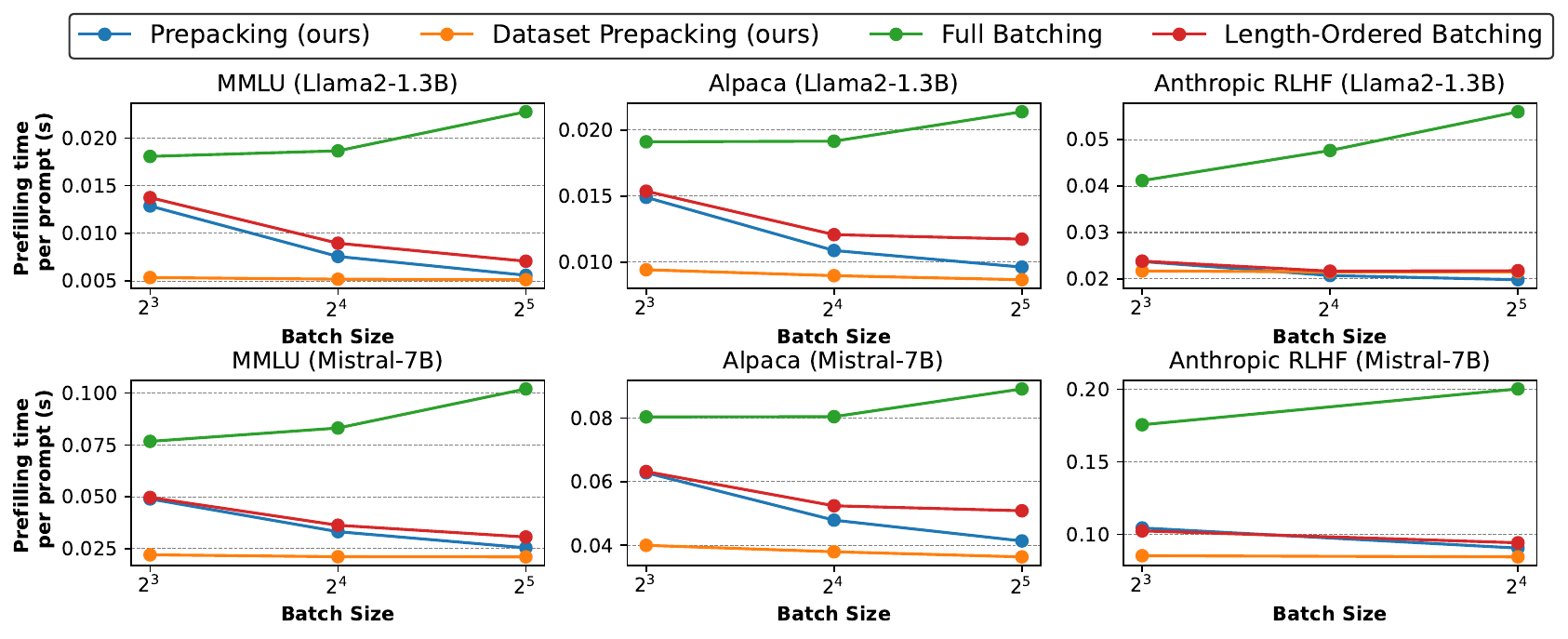}
    \caption{Comparison of prefilling time per prompt. The Dataset \name{} and Length-Ordered Batching benefit from access to the entire dataset, in contrast to Batch \name{} and Full Batching, which operate on a per-batch basis. Dataset \name{} further minimizes prefilling latency through packing when provided with full dataset access. Results are averaged over 5 runs.}
    \label{datasetpack}
    \end{center}
\end{figure}

In the previous experiments, we apply \name{} on randomly sampled batches from each dataset. However, this assumes the inability to control the contents of each batch. Given the ability to determine batches, a method to padding inefficiency would be to sort the dataset by length and batch accordingly. We refer to this baseline as \textit{Length-Ordered Batching}. Alternatively, we can create batches after performing \name{} on the dataset as a whole and apply \name{}, i.e. \textit{Dataset \cname{}}. We find that even in this scenario, where one might expect length-ordered batching to have a near optimal runtime by reducing the number of pad tokens, we observe \name{} still exhibits improvements as shown in Figure~\ref{datasetpack}, where we compare the prefilling time per prompt.


\subsection{How does the performance gain scale with characteristics of lengths within a batch?} 
\label{sec:regerssion}
\begin{figure}[h]
\begin{center}
    \includegraphics[width=0.85\textwidth]{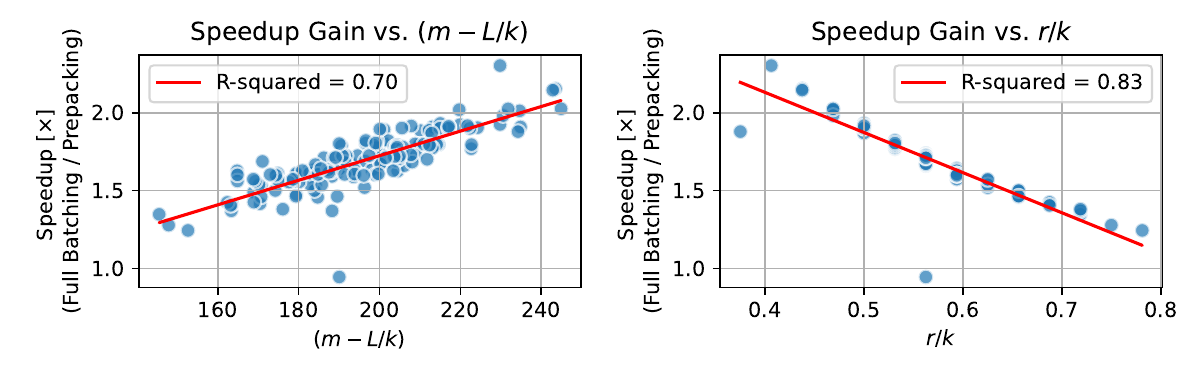}
    \caption{We show that speedup can be predicted from two dataset statistics: Max Absolute Deviation ($m-L/n$) and Batch Size Reduction ($r/n$). We perform a linear regression and observe a high correlation between these statistics and observed speedup.}
    \label{dataset_statistics}
    \end{center}
\end{figure}

Previously in Section \ref{runtime}, we find the runtime of full batching is $\mathcal{O}(km^2)$. \cname{} is $\mathcal{O}(rm^2)$, where $k$ is the original batch size, $r$ is the batch size after \name{}, and $m$ is the maximum prompt length. Therefore, we can estimate the speedup as a function of $r/k$ (Batch Size Reduction). Because in practice it is difficult to predict $r$ from the dataset statistics alone, we can also estimate the speedup as a function of $m - L/k$ (Max Absolute Deviation), which is how much the maximum length of a batch deviates from the  mean length. We conduct the analysis on 5000 synthetic prompts with lengths uniformly distributed from 1 to 512, using the Llama2 1.3B model with batch size of 32. As can be seen in Figure \ref{dataset_statistics}, these metrics can predict the speedup obtained by using \name{} over full batching. We show more plots with different model and batch size in Appendix~\ref{sec:statisticsplot_more}.

\subsection{\cname{} for Generation}
Beyong prefilling, the concept of packing holds great promise for LLM generation. Padding is also a problem for generation because when new tokens are generated, their queries must be dotted with cached keys and values, which inevitably contains padding. With a similar technique to \name{}, we can bin-pack the KV caches of larger batch sizes with varying lengths into smaller batch sizes at generation time, saving memory that would be otherwise wasted on padding. We present preliminary results using a vanilla implementation of generation with \name{} and Llama2 1.3B on a toy batch of size 10, where 9 of the prompts are of length 1 and one prompt is of length 1000. We show the memory usage and generation time (excluding prefilling) for the next 20 tokens in Table~\ref{tab:generation}. \cname{} uses 56\% less peak GPU memory and offers 35\% faster generation times. While the use of packing for generation is a promising direction for future work, it is beyond the scope of this paper.

\begin{table}[h]
\centering
\scalebox{0.88}{
\begin{tabular}{c|c|c|c|c}
Generation Method & \multicolumn{1}{c|}{\makecell{Peak GPU\\ Memory (GB)}} & \multicolumn{1}{c|}{\makecell{Memory\\ Savings vs.\\ Full Batching (\%)}} & \multicolumn{1}{c|}{Generation Time (s)} & \multicolumn{1}{c}{\makecell{Time\\ Savings vs.\\ Full Batching (\%)}}\\
\hline
Full Batching & 15.908 & - & 1.253 $\pm$ 0.007 & - \\
\hline
Prepacking & 6.940 & 56.374 & 0.812 $\pm$ 0.008 & 35.196 \\
\end{tabular}
}
\caption{Comparison of peak GPU memory usage and generation time for the next 20 tokens on a toy example. \cname{} reduces peak GPU memory usage by 56\% and offers 35\% faster generation times compared to Full Batching. Results are averaged over 20 runs and standard deviations of generation time are reported.}
\label{tab:generation}
\end{table}


\section{Related Works}

\subsection{Accelerating Inference}
Many advancements in accelerating LLM inference make architectural modifications that tradeoff quality with inference latency.
These approaches include exploiting contextual sparsity~\citep{liu2023deja}, multiple decoding heads ~\citep{cai2024medusa}, model quantization ~\citep{xiao2023smoothquant}, and improved decoding algorithms such as speculative decoding which augments a base model with an ``approximation model" ~\citep{leviathan2023fast}. Another active area of research is speeding up inference by improving low-level compute scheduling ~\citep{aminabadi2022deepspeed, sheng2023flexgen}. Our approach for improving LLM throughput differs from the aforementioned techniques because: (1) it does not require any architectural changes; (2) it can be fully implemented in PyTorch and is agnostic to the underlying hardware and cloud platforms.

\subsection{LLM Serving}
A relevant line of work takes a networking perspective on LLMs, in which a model must be ``served" to clients that make requests. The core problem LLM serving addresses is the scheduling of inference, creating dynamic schedulers that optimize for throughput and latency. FasterTransformer \citep{githubGitHubNVIDIAFasterTransformer} increases decoding throughput but schedules at the request-level. To address this, Orca ~\citep{yu2022orca} proposes iteration-level scheduling which processes requests at finer granularity than a batch. PagedAttention in vLLM ~\citep{kwon2023efficient} reduces KV-cache memory fragmentation with techniques inspired by virtual memory with paging. More recent  and concurrent works such as Sarathi-Serve ~\citep{agrawal2024taming} and DistServe ~\citep{zhong2024distserve} optimize a trade-off involving pre-filling and decoding. In our work, we specifically target pre-filling only. As such, our work directly improves TTFT and is complementary to other works that seek to improve decoding efficiency and throughput while minimizing stalling.

\section{Conclusion}
We proposed \name{}, a simple and effective approach to optimize the prefilling computation for LLMs during inference. Our evaluation on typical datasets with varying prompt lengths demonstrates significant speedups compared to standard prefilling computation in Huggingface's implementation. As language models continue to scale and support longer context lengths, addressing the inefficiencies associated with prefilling computation becomes crucial for optimizing inference speed and computational resource allocation. Prepacking provides a promising solution to this challenge, enabling more efficient inference for prompts with varying lengths.
In the future, we plan to explore more complex decoding strategies post-prefilling that also incorporate bin packing for further increase in throughput.

\newpage
\bibliography{colm2024_conference}

\begin{thebibliography}{33}
\providecommand{\natexlab}[1]{#1}
\providecommand{\url}[1]{\texttt{#1}}
\expandafter\ifx\csname urlstyle\endcsname\relax
  \providecommand{\doi}[1]{doi: #1}\else
  \providecommand{\doi}{doi: \begingroup \urlstyle{rm}\Url}\fi

\bibitem[Achiam et~al.(2023)Achiam, Adler, Agarwal, Ahmad, Akkaya, Aleman, Almeida, Altenschmidt, Altman, Anadkat, et~al.]{achiam2023gpt}
Josh Achiam, Steven Adler, Sandhini Agarwal, Lama Ahmad, Ilge Akkaya, Florencia~Leoni Aleman, Diogo Almeida, Janko Altenschmidt, Sam Altman, Shyamal Anadkat, et~al.
\newblock Gpt-4 technical report.
\newblock \emph{arXiv preprint arXiv:2303.08774}, 2023.

\bibitem[Agrawal et~al.(2024)Agrawal, Kedia, Panwar, Mohan, Kwatra, Gulavani, Tumanov, and Ramjee]{agrawal2024taming}
Amey Agrawal, Nitin Kedia, Ashish Panwar, Jayashree Mohan, Nipun Kwatra, Bhargav~S Gulavani, Alexey Tumanov, and Ramachandran Ramjee.
\newblock Taming throughput-latency tradeoff in llm inference with sarathi-serve.
\newblock \emph{arXiv preprint arXiv:2403.02310}, 2024.

\bibitem[Aminabadi et~al.(2022)Aminabadi, Rajbhandari, Awan, Li, Li, Zheng, Ruwase, Smith, Zhang, Rasley, et~al.]{aminabadi2022deepspeed}
Reza~Yazdani Aminabadi, Samyam Rajbhandari, Ammar~Ahmad Awan, Cheng Li, Du~Li, Elton Zheng, Olatunji Ruwase, Shaden Smith, Minjia Zhang, Jeff Rasley, et~al.
\newblock Deepspeed-inference: enabling efficient inference of transformer models at unprecedented scale.
\newblock In \emph{SC22: International Conference for High Performance Computing, Networking, Storage and Analysis}, pp.\  1--15. IEEE, 2022.

\bibitem[Bai et~al.(2022)Bai, Jones, Ndousse, Askell, Chen, DasSarma, Drain, Fort, Ganguli, Henighan, et~al.]{bai2022training}
Yuntao Bai, Andy Jones, Kamal Ndousse, Amanda Askell, Anna Chen, Nova DasSarma, Dawn Drain, Stanislav Fort, Deep Ganguli, Tom Henighan, et~al.
\newblock Training a helpful and harmless assistant with reinforcement learning from human feedback.
\newblock \emph{arXiv preprint arXiv:2204.05862}, 2022.

\bibitem[Buljuba{\v{s}}i{\'c} \& Vasquez(2016)Buljuba{\v{s}}i{\'c} and Vasquez]{buljubavsic2016consistent}
Mirsad Buljuba{\v{s}}i{\'c} and Michel Vasquez.
\newblock Consistent neighborhood search for one-dimensional bin packing and two-dimensional vector packing.
\newblock \emph{Computers \& Operations Research}, 76:\penalty0 12--21, 2016.

\bibitem[Cai et~al.(2024)Cai, Li, Geng, Peng, Lee, Chen, and Dao]{cai2024medusa}
Tianle Cai, Yuhong Li, Zhengyang Geng, Hongwu Peng, Jason~D. Lee, Deming Chen, and Tri Dao.
\newblock Medusa: Simple llm inference acceleration framework with multiple decoding heads, 2024.

\bibitem[Eloundou et~al.(2023)Eloundou, Manning, Mishkin, and Rock]{eloundou2023gpts}
Tyna Eloundou, Sam Manning, Pamela Mishkin, and Daniel Rock.
\newblock Gpts are gpts: An early look at the labor market impact potential of large language models.
\newblock \emph{arXiv preprint arXiv:2303.10130}, 2023.

\bibitem[Garey \& Johnson(1979)Garey and Johnson]{garey1979computers}
MR~Garey and DS~Johnson.
\newblock Computers and intractability a guide to the theory of np-completeness new york freeman and co.
\newblock 1979.

\bibitem[Gliwa et~al.(2019)Gliwa, Mochol, Biesek, and Wawer]{gliwa-etal-2019-samsum}
Bogdan Gliwa, Iwona Mochol, Maciej Biesek, and Aleksander Wawer.
\newblock {SAMS}um corpus: A human-annotated dialogue dataset for abstractive summarization.
\newblock In \emph{Proceedings of the 2nd Workshop on New Frontiers in Summarization}, pp.\  70--79, Hong Kong, China, November 2019. Association for Computational Linguistics.
\newblock \doi{10.18653/v1/D19-5409}.
\newblock URL \url{https://www.aclweb.org/anthology/D19-5409}.

\bibitem[Hendrycks et~al.(2021{\natexlab{a}})Hendrycks, Burns, Basart, Critch, Li, Song, and Steinhardt]{hendrycks2021ethics}
Dan Hendrycks, Collin Burns, Steven Basart, Andrew Critch, Jerry Li, Dawn Song, and Jacob Steinhardt.
\newblock Aligning ai with shared human values.
\newblock \emph{Proceedings of the International Conference on Learning Representations (ICLR)}, 2021{\natexlab{a}}.

\bibitem[Hendrycks et~al.(2021{\natexlab{b}})Hendrycks, Burns, Basart, Zou, Mazeika, Song, and Steinhardt]{hendryckstest2021}
Dan Hendrycks, Collin Burns, Steven Basart, Andy Zou, Mantas Mazeika, Dawn Song, and Jacob Steinhardt.
\newblock Measuring massive multitask language understanding.
\newblock \emph{Proceedings of the International Conference on Learning Representations (ICLR)}, 2021{\natexlab{b}}.

\bibitem[Jiang et~al.(2023)Jiang, Sablayrolles, Mensch, Bamford, Chaplot, Casas, Bressand, Lengyel, Lample, Saulnier, et~al.]{jiang2023mistral}
Albert~Q Jiang, Alexandre Sablayrolles, Arthur Mensch, Chris Bamford, Devendra~Singh Chaplot, Diego de~las Casas, Florian Bressand, Gianna Lengyel, Guillaume Lample, Lucile Saulnier, et~al.
\newblock Mistral 7b.
\newblock \emph{arXiv preprint arXiv:2310.06825}, 2023.

\bibitem[Kwon et~al.(2023)Kwon, Li, Zhuang, Sheng, Zheng, Yu, Gonzalez, Zhang, and Stoica]{kwon2023efficient}
Woosuk Kwon, Zhuohan Li, Siyuan Zhuang, Ying Sheng, Lianmin Zheng, Cody~Hao Yu, Joseph~E. Gonzalez, Hao Zhang, and Ion Stoica.
\newblock Efficient memory management for large language model serving with pagedattention, 2023.

\bibitem[Leviathan et~al.(2023)Leviathan, Kalman, and Matias]{leviathan2023fast}
Yaniv Leviathan, Matan Kalman, and Yossi Matias.
\newblock Fast inference from transformers via speculative decoding.
\newblock In \emph{International Conference on Machine Learning}, pp.\  19274--19286. PMLR, 2023.

\bibitem[Liu et~al.(2023)Liu, Wang, Dao, Zhou, Yuan, Song, Shrivastava, Zhang, Tian, Re, et~al.]{liu2023deja}
Zichang Liu, Jue Wang, Tri Dao, Tianyi Zhou, Binhang Yuan, Zhao Song, Anshumali Shrivastava, Ce~Zhang, Yuandong Tian, Christopher Re, et~al.
\newblock Deja vu: Contextual sparsity for efficient llms at inference time.
\newblock In \emph{International Conference on Machine Learning}, pp.\  22137--22176. PMLR, 2023.

\bibitem[Maier(2021)]{githubGitHubBenmaierbinpacking}
Ben Maier.
\newblock {G}it{H}ub - benmaier/binpacking: {D}istribution of weighted items to bins (either a fixed number of bins or a fixed number of volume per bin). {D}ata may be in form of list, dictionary, list of tuples or csv-file. --- github.com.
\newblock \url{https://github.com/benmaier/binpacking}, 2021.
\newblock [Accessed 30-03-2024].

\bibitem[Merity et~al.(2016)Merity, Xiong, Bradbury, and Socher]{merity2016pointer}
Stephen Merity, Caiming Xiong, James Bradbury, and Richard Socher.
\newblock Pointer sentinel mixture models, 2016.

\bibitem[Miao et~al.(2023)Miao, Oliaro, Zhang, Cheng, Jin, Chen, and Jia]{miao2023towards}
Xupeng Miao, Gabriele Oliaro, Zhihao Zhang, Xinhao Cheng, Hongyi Jin, Tianqi Chen, and Zhihao Jia.
\newblock Towards efficient generative large language model serving: A survey from algorithms to systems.
\newblock \emph{arXiv preprint arXiv:2312.15234}, 2023.

\bibitem[Naseer et~al.(2021)Naseer, Ranasinghe, Khan, Hayat, Shahbaz~Khan, and Yang]{naseer2021intriguing}
Muhammad~Muzammal Naseer, Kanchana Ranasinghe, Salman~H Khan, Munawar Hayat, Fahad Shahbaz~Khan, and Ming-Hsuan Yang.
\newblock Intriguing properties of vision transformers.
\newblock \emph{Advances in Neural Information Processing Systems}, 34:\penalty0 23296--23308, 2021.

\bibitem[NVIDIA(2021)]{githubGitHubNVIDIAFasterTransformer}
NVIDIA.
\newblock {G}it{H}ub - {N}{V}{I}{D}{I}{A}/{F}aster{T}ransformer: {T}ransformer related optimization, including {B}{E}{R}{T}, {G}{P}{T} --- github.com.
\newblock \url{https://github.com/NVIDIA/FasterTransformer}, 2021.
\newblock [Accessed 29-03-2024].

\bibitem[Pope et~al.(2023)Pope, Douglas, Chowdhery, Devlin, Bradbury, Heek, Xiao, Agrawal, and Dean]{pope2023efficiently}
Reiner Pope, Sholto Douglas, Aakanksha Chowdhery, Jacob Devlin, James Bradbury, Jonathan Heek, Kefan Xiao, Shivani Agrawal, and Jeff Dean.
\newblock Efficiently scaling transformer inference.
\newblock \emph{Proceedings of Machine Learning and Systems}, 5, 2023.

\bibitem[Radford et~al.(2019)Radford, Wu, Child, Luan, Amodei, Sutskever, et~al.]{radford2019language}
Alec Radford, Jeffrey Wu, Rewon Child, David Luan, Dario Amodei, Ilya Sutskever, et~al.
\newblock Language models are unsupervised multitask learners.
\newblock \emph{OpenAI blog}, 1\penalty0 (8):\penalty0 9, 2019.

\bibitem[Reid et~al.(2024)Reid, Savinov, Teplyashin, Lepikhin, Lillicrap, baptiste Alayrac, Soricut, Lazaridou, Firat, Schrittwieser, Antonoglou, Anil, Borgeaud, Dai, Millican, Dyer, Glaese, Sottiaux, Lee, Viola, Reynolds, Xu, Molloy, Chen, Isard, Barham, Hennigan, McIlroy, Johnson, Schalkwyk, Collins, Rutherford, Moreira, Ayoub, Goel, Meyer, Thornton, Yang, Michalewski, Abbas, Schucher, Anand, Ives, Keeling, Lenc, Haykal, Shakeri, Shyam, Chowdhery, Ring, Spencer, Sezener, Vilnis, Chang, Morioka, Tucker, Zheng, Woodman, Attaluri, Kocisky, Eltyshev, Chen, Chung, Selo, Brahma, Georgiev, Slone, Zhu, Lottes, Qiao, Caine, Riedel, Tomala, Chadwick, Love, Choy, Mittal, Houlsby, Tang, Lamm, Bai, Zhang, He, Cheng, Humphreys, Li, Brin, Cassirer, Miao, Zilka, Tobin, Xu, Proleev, Sohn, Magni, Hendricks, Gao, Ontañón, Bunyan, Byrd, Sharma, Zhang, Pinto, Sinha, Mehta, Jia, Caelles, Webson, Morris, Roelofs, Ding, Strudel, Xiong, Ritter, Dehghani, Chaabouni, Karmarkar, Lai, Mentzer, Xu, Li, Zhang, Paine, Goldin, Neyshabur,
  Baumli, Levskaya, Laskin, Jia, Rae, Xiao, He, Giordano, Yagati, Lespiau, Natsev, Ganapathy, Liu, Martins, Chen, Xu, Barnes, May, Vezer, Oh, Franko, Bridgers, Zhao, Wu, Mustafa, Sechrist, Parisotto, Pillai, Larkin, Gu, Sorokin, Krikun, Guseynov, Landon, Datta, Pritzel, Thacker, Yang, Hui, Hauth, Yeh, Barker, Mao-Jones, Austin, Sheahan, Schuh, Svensson, Jain, Ramasesh, Briukhov, Chung, von Glehn, Butterfield, Jhakra, Wiethoff, Frye, Grimstad, Changpinyo, Lan, Bortsova, Wu, Voigtlaender, Sainath, Smith, Hawkins, Cao, Besley, Srinivasan, Omernick, Gaffney, Surita, Burnell, Damoc, Ahn, Brock, Pajarskas, Petrushkina, Noury, Blanco, Swersky, Ahuja, Avrahami, Misra, de~Liedekerke, Iinuma, Polozov, York, van~den Driessche, Michel, Chiu, Blevins, Gleicher, Recasens, Rrustemi, Gribovskaya, Roy, Gworek, Arnold, Lee, Lee-Thorp, Maggioni, Piqueras, Badola, Vikram, Gonzalez, Baddepudi, Senter, Devlin, Qin, Azzam, Trebacz, Polacek, Krishnakumar, yiin Chang, Tung, Penchev, Joshi, Olszewska, Muir, Wirth, Hartman, Newlan,
  Kashem, Bolina, Dabir, van Amersfoort, Ahmed, Cobon-Kerr, Kamath, Hrafnkelsson, Hou, Mackinnon, Frechette, Noland, Si, Taropa, Li, Crone, Gulati, Cevey, Adler, Ma, Silver, Tokumine, Powell, Lee, Chang, Hassan, Mincu, Yang, Levine, Brennan, Wang, Hodkinson, Zhao, Lipschultz, Pope, Chang, Li, Shafey, Paganini, Douglas, Bohnet, Pardo, Odoom, Rosca, dos Santos, Soparkar, Guez, Hudson, Hansen, Asawaroengchai, Addanki, Yu, Stokowiec, Khan, Gilmer, Lee, Bostock, Rong, Caton, Pejman, Pavetic, Brown, Sharma, Lučić, Samuel, Djolonga, Mandhane, Sjösund, Buchatskaya, White, Clay, Jiang, Lim, Hemsley, Labanowski, Cao, Steiner, Hashemi, Austin, Gergely, Blyth, Stanton, Shivakumar, Siddhant, Andreassen, Araya, Sethi, Shivanna, Hand, Bapna, Khodaei, Miech, Tanzer, Swing, Thakoor, Pan, Nado, Winkler, Yu, Saleh, Maggiore, Barr, Giang, Kagohara, Danihelka, Marathe, Feinberg, Elhawaty, Ghelani, Horgan, Miller, Walker, Tanburn, Tariq, Shrivastava, Xia, Chiu, Ashwood, Baatarsukh, Samangooei, Alcober, Stjerngren, Komarek,
  Tsihlas, Boral, Comanescu, Chen, Liu, Bloxwich, Chen, Sun, Feng, Mauger, Dotiwalla, Hellendoorn, Sharman, Zheng, Haridasan, Barth-Maron, Swanson, Rogozińska, Andreev, Rubenstein, Sang, Hurt, Elsayed, Wang, Lacey, Ilić, Zhao, Aroyo, Iwuanyanwu, Nikolaev, Lakshminarayanan, Jazayeri, Kaufman, Varadarajan, Tekur, Fritz, Khalman, Reitter, Dasgupta, Sarcar, Ornduff, Snaider, Huot, Jia, Kemp, Trdin, Vijayakumar, Kim, Angermueller, Lao, Liu, Zhang, Engel, Greene, White, Austin, Taylor, Ashraf, Liu, Georgaki, Cai, Kulizhskaya, Goenka, Saeta, Vodrahalli, Frank, de~Cesare, Robenek, Richardson, Alnahlawi, Yew, Ponnapalli, Tagliasacchi, Korchemniy, Kim, Li, Rosgen, Ashwood, Levin, Wiesner, Banzal, Srinivasan, Yu, Çağlar Ünlü, Reid, Tung, Finchelstein, Kumar, Elisseeff, Huang, Zhang, Zhu, Aguilar, Giménez, Xia, Dousse, Gierke, Yeganeh, Yates, Jalan, Li, Latorre-Chimoto, Nguyen, Durden, Kallakuri, Liu, Johnson, Tsai, Talbert, Liu, Neitz, Elkind, Selvi, Jasarevic, Soares, Cui, Wang, Wang, Ye, Kallarackal, Loher,
  Lam, Broder, Holtmann-Rice, Martin, Ramadhana, Toyama, Shukla, Basu, Mohan, Fernando, Fiedel, Paterson, Li, Garg, Park, Choi, Wu, Singh, Zhang, Globerson, Yu, Carpenter, de~Chaumont~Quitry, Radebaugh, Lin, Tudor, Shroff, Garmon, Du, Vats, Lu, Iqbal, Yakubovich, Tripuraneni, Manyika, Qureshi, Hua, Ngani, Raad, Forbes, Bulanova, Stanway, Sundararajan, Ungureanu, Bishop, Li, Venkatraman, Li, Thornton, Scellato, Gupta, Wang, Tenney, Wu, Shenoy, Carvajal, Wright, Bariach, Xiao, Hawkins, Dalmia, Farabet, Valenzuela, Yuan, Welty, Agarwal, Chen, Kim, Hulse, Dukkipati, Paszke, Bolt, Davoodi, Choo, Beattie, Prendki, Vashisht, Santamaria-Fernandez, Cobo, Wilkiewicz, Madras, Elqursh, Uy, Ramirez, Harvey, Liechty, Zen, Seibert, Hu, Elhawaty, Khorlin, Le, Aharoni, Li, Wang, Kumar, Lince, Casagrande, Hoover, Badawy, Soergel, Vnukov, Miecnikowski, Simsa, Koop, Kumar, Sellam, Vlasic, Daruki, Shabat, Zhang, Su, Zhang, Liu, Sun, Palmer, Ghaffarkhah, Xiong, Cotruta, Fink, Dixon, Sreevatsa, Goedeckemeyer, Dimitriev, Jafari,
  Crocker, FitzGerald, Kumar, Ghemawat, Philips, Liu, Liang, Sterneck, Repina, Wu, Knight, Georgiev, Lee, Askham, Chakladar, Louis, Crous, Cate, Petrova, Quinn, Owusu-Afriyie, Singhal, Wei, Kim, Vincent, Nasr, Choquette-Choo, Tojo, Lu, de~Las~Casas, Cheng, Bolukbasi, Lee, Fatehi, Ananthanarayanan, Patel, Kaed, Li, Sygnowski, Belle, Chen, Konzelmann, Põder, Garg, Koverkathu, Brown, Dyer, Liu, Nova, Xu, Petrov, Hassabis, Kavukcuoglu, Dean, and Vinyals]{reid2024gemini}
Machel Reid, Nikolay Savinov, Denis Teplyashin, Dmitry Lepikhin, Timothy Lillicrap, Jean baptiste Alayrac, Radu Soricut, Angeliki Lazaridou, Orhan Firat, Julian Schrittwieser, Ioannis Antonoglou, Rohan Anil, Sebastian Borgeaud, Andrew Dai, Katie Millican, Ethan Dyer, Mia Glaese, Thibault Sottiaux, Benjamin Lee, Fabio Viola, Malcolm Reynolds, Yuanzhong Xu, James Molloy, Jilin Chen, Michael Isard, Paul Barham, Tom Hennigan, Ross McIlroy, Melvin Johnson, Johan Schalkwyk, Eli Collins, Eliza Rutherford, Erica Moreira, Kareem Ayoub, Megha Goel, Clemens Meyer, Gregory Thornton, Zhen Yang, Henryk Michalewski, Zaheer Abbas, Nathan Schucher, Ankesh Anand, Richard Ives, James Keeling, Karel Lenc, Salem Haykal, Siamak Shakeri, Pranav Shyam, Aakanksha Chowdhery, Roman Ring, Stephen Spencer, Eren Sezener, Luke Vilnis, Oscar Chang, Nobuyuki Morioka, George Tucker, Ce~Zheng, Oliver Woodman, Nithya Attaluri, Tomas Kocisky, Evgenii Eltyshev, Xi~Chen, Timothy Chung, Vittorio Selo, Siddhartha Brahma, Petko Georgiev, Ambrose
  Slone, Zhenkai Zhu, James Lottes, Siyuan Qiao, Ben Caine, Sebastian Riedel, Alex Tomala, Martin Chadwick, Juliette Love, Peter Choy, Sid Mittal, Neil Houlsby, Yunhao Tang, Matthew Lamm, Libin Bai, Qiao Zhang, Luheng He, Yong Cheng, Peter Humphreys, Yujia Li, Sergey Brin, Albin Cassirer, Yingjie Miao, Lukas Zilka, Taylor Tobin, Kelvin Xu, Lev Proleev, Daniel Sohn, Alberto Magni, Lisa~Anne Hendricks, Isabel Gao, Santiago Ontañón, Oskar Bunyan, Nathan Byrd, Abhanshu Sharma, Biao Zhang, Mario Pinto, Rishika Sinha, Harsh Mehta, Dawei Jia, Sergi Caelles, Albert Webson, Alex Morris, Becca Roelofs, Yifan Ding, Robin Strudel, Xuehan Xiong, Marvin Ritter, Mostafa Dehghani, Rahma Chaabouni, Abhijit Karmarkar, Guangda Lai, Fabian Mentzer, Bibo Xu, YaGuang Li, Yujing Zhang, Tom~Le Paine, Alex Goldin, Behnam Neyshabur, Kate Baumli, Anselm Levskaya, Michael Laskin, Wenhao Jia, Jack~W. Rae, Kefan Xiao, Antoine He, Skye Giordano, Lakshman Yagati, Jean-Baptiste Lespiau, Paul Natsev, Sanjay Ganapathy, Fangyu Liu, Danilo
  Martins, Nanxin Chen, Yunhan Xu, Megan Barnes, Rhys May, Arpi Vezer, Junhyuk Oh, Ken Franko, Sophie Bridgers, Ruizhe Zhao, Boxi Wu, Basil Mustafa, Sean Sechrist, Emilio Parisotto, Thanumalayan~Sankaranarayana Pillai, Chris Larkin, Chenjie Gu, Christina Sorokin, Maxim Krikun, Alexey Guseynov, Jessica Landon, Romina Datta, Alexander Pritzel, Phoebe Thacker, Fan Yang, Kevin Hui, Anja Hauth, Chih-Kuan Yeh, David Barker, Justin Mao-Jones, Sophia Austin, Hannah Sheahan, Parker Schuh, James Svensson, Rohan Jain, Vinay Ramasesh, Anton Briukhov, Da-Woon Chung, Tamara von Glehn, Christina Butterfield, Priya Jhakra, Matthew Wiethoff, Justin Frye, Jordan Grimstad, Beer Changpinyo, Charline~Le Lan, Anna Bortsova, Yonghui Wu, Paul Voigtlaender, Tara Sainath, Charlotte Smith, Will Hawkins, Kris Cao, James Besley, Srivatsan Srinivasan, Mark Omernick, Colin Gaffney, Gabriela Surita, Ryan Burnell, Bogdan Damoc, Junwhan Ahn, Andrew Brock, Mantas Pajarskas, Anastasia Petrushkina, Seb Noury, Lorenzo Blanco, Kevin Swersky, Arun
  Ahuja, Thi Avrahami, Vedant Misra, Raoul de~Liedekerke, Mariko Iinuma, Alex Polozov, Sarah York, George van~den Driessche, Paul Michel, Justin Chiu, Rory Blevins, Zach Gleicher, Adrià Recasens, Alban Rrustemi, Elena Gribovskaya, Aurko Roy, Wiktor Gworek, Séb Arnold, Lisa Lee, James Lee-Thorp, Marcello Maggioni, Enrique Piqueras, Kartikeya Badola, Sharad Vikram, Lucas Gonzalez, Anirudh Baddepudi, Evan Senter, Jacob Devlin, James Qin, Michael Azzam, Maja Trebacz, Martin Polacek, Kashyap Krishnakumar, Shuo yiin Chang, Matthew Tung, Ivo Penchev, Rishabh Joshi, Kate Olszewska, Carrie Muir, Mateo Wirth, Ale~Jakse Hartman, Josh Newlan, Sheleem Kashem, Vijay Bolina, Elahe Dabir, Joost van Amersfoort, Zafarali Ahmed, James Cobon-Kerr, Aishwarya Kamath, Arnar~Mar Hrafnkelsson, Le~Hou, Ian Mackinnon, Alexandre Frechette, Eric Noland, Xiance Si, Emanuel Taropa, Dong Li, Phil Crone, Anmol Gulati, Sébastien Cevey, Jonas Adler, Ada Ma, David Silver, Simon Tokumine, Richard Powell, Stephan Lee, Michael Chang, Samer
  Hassan, Diana Mincu, Antoine Yang, Nir Levine, Jenny Brennan, Mingqiu Wang, Sarah Hodkinson, Jeffrey Zhao, Josh Lipschultz, Aedan Pope, Michael~B. Chang, Cheng Li, Laurent~El Shafey, Michela Paganini, Sholto Douglas, Bernd Bohnet, Fabio Pardo, Seth Odoom, Mihaela Rosca, Cicero~Nogueira dos Santos, Kedar Soparkar, Arthur Guez, Tom Hudson, Steven Hansen, Chulayuth Asawaroengchai, Ravi Addanki, Tianhe Yu, Wojciech Stokowiec, Mina Khan, Justin Gilmer, Jaehoon Lee, Carrie~Grimes Bostock, Keran Rong, Jonathan Caton, Pedram Pejman, Filip Pavetic, Geoff Brown, Vivek Sharma, Mario Lučić, Rajkumar Samuel, Josip Djolonga, Amol Mandhane, Lars~Lowe Sjösund, Elena Buchatskaya, Elspeth White, Natalie Clay, Jiepu Jiang, Hyeontaek Lim, Ross Hemsley, Jane Labanowski, Nicola~De Cao, David Steiner, Sayed~Hadi Hashemi, Jacob Austin, Anita Gergely, Tim Blyth, Joe Stanton, Kaushik Shivakumar, Aditya Siddhant, Anders Andreassen, Carlos Araya, Nikhil Sethi, Rakesh Shivanna, Steven Hand, Ankur Bapna, Ali Khodaei, Antoine Miech,
  Garrett Tanzer, Andy Swing, Shantanu Thakoor, Zhufeng Pan, Zachary Nado, Stephanie Winkler, Dian Yu, Mohammad Saleh, Loren Maggiore, Iain Barr, Minh Giang, Thais Kagohara, Ivo Danihelka, Amit Marathe, Vladimir Feinberg, Mohamed Elhawaty, Nimesh Ghelani, Dan Horgan, Helen Miller, Lexi Walker, Richard Tanburn, Mukarram Tariq, Disha Shrivastava, Fei Xia, Chung-Cheng Chiu, Zoe Ashwood, Khuslen Baatarsukh, Sina Samangooei, Fred Alcober, Axel Stjerngren, Paul Komarek, Katerina Tsihlas, Anudhyan Boral, Ramona Comanescu, Jeremy Chen, Ruibo Liu, Dawn Bloxwich, Charlie Chen, Yanhua Sun, Fangxiaoyu Feng, Matthew Mauger, Xerxes Dotiwalla, Vincent Hellendoorn, Michael Sharman, Ivy Zheng, Krishna Haridasan, Gabe Barth-Maron, Craig Swanson, Dominika Rogozińska, Alek Andreev, Paul~Kishan Rubenstein, Ruoxin Sang, Dan Hurt, Gamaleldin Elsayed, Renshen Wang, Dave Lacey, Anastasija Ilić, Yao Zhao, Lora Aroyo, Chimezie Iwuanyanwu, Vitaly Nikolaev, Balaji Lakshminarayanan, Sadegh Jazayeri, Raphaël~Lopez Kaufman, Mani
  Varadarajan, Chetan Tekur, Doug Fritz, Misha Khalman, David Reitter, Kingshuk Dasgupta, Shourya Sarcar, Tina Ornduff, Javier Snaider, Fantine Huot, Johnson Jia, Rupert Kemp, Nejc Trdin, Anitha Vijayakumar, Lucy Kim, Christof Angermueller, Li~Lao, Tianqi Liu, Haibin Zhang, David Engel, Somer Greene, Anaïs White, Jessica Austin, Lilly Taylor, Shereen Ashraf, Dangyi Liu, Maria Georgaki, Irene Cai, Yana Kulizhskaya, Sonam Goenka, Brennan Saeta, Kiran Vodrahalli, Christian Frank, Dario de~Cesare, Brona Robenek, Harry Richardson, Mahmoud Alnahlawi, Christopher Yew, Priya Ponnapalli, Marco Tagliasacchi, Alex Korchemniy, Yelin Kim, Dinghua Li, Bill Rosgen, Zoe Ashwood, Kyle Levin, Jeremy Wiesner, Praseem Banzal, Praveen Srinivasan, Hongkun Yu, Çağlar Ünlü, David Reid, Zora Tung, Daniel Finchelstein, Ravin Kumar, Andre Elisseeff, Jin Huang, Ming Zhang, Rui Zhu, Ricardo Aguilar, Mai Giménez, Jiawei Xia, Olivier Dousse, Willi Gierke, Soheil~Hassas Yeganeh, Damion Yates, Komal Jalan, Lu~Li, Eri Latorre-Chimoto,
  Duc~Dung Nguyen, Ken Durden, Praveen Kallakuri, Yaxin Liu, Matthew Johnson, Tomy Tsai, Alice Talbert, Jasmine Liu, Alexander Neitz, Chen Elkind, Marco Selvi, Mimi Jasarevic, Livio~Baldini Soares, Albert Cui, Pidong Wang, Alek~Wenjiao Wang, Xinyu Ye, Krystal Kallarackal, Lucia Loher, Hoi Lam, Josef Broder, Dan Holtmann-Rice, Nina Martin, Bramandia Ramadhana, Daniel Toyama, Mrinal Shukla, Sujoy Basu, Abhi Mohan, Nick Fernando, Noah Fiedel, Kim Paterson, Hui Li, Ankush Garg, Jane Park, DongHyun Choi, Diane Wu, Sankalp Singh, Zhishuai Zhang, Amir Globerson, Lily Yu, John Carpenter, Félix de~Chaumont~Quitry, Carey Radebaugh, Chu-Cheng Lin, Alex Tudor, Prakash Shroff, Drew Garmon, Dayou Du, Neera Vats, Han Lu, Shariq Iqbal, Alex Yakubovich, Nilesh Tripuraneni, James Manyika, Haroon Qureshi, Nan Hua, Christel Ngani, Maria~Abi Raad, Hannah Forbes, Anna Bulanova, Jeff Stanway, Mukund Sundararajan, Victor Ungureanu, Colton Bishop, Yunjie Li, Balaji Venkatraman, Bo~Li, Chloe Thornton, Salvatore Scellato, Nishesh
  Gupta, Yicheng Wang, Ian Tenney, Xihui Wu, Ashish Shenoy, Gabriel Carvajal, Diana~Gage Wright, Ben Bariach, Zhuyun Xiao, Peter Hawkins, Sid Dalmia, Clement Farabet, Pedro Valenzuela, Quan Yuan, Chris Welty, Ananth Agarwal, Mia Chen, Wooyeol Kim, Brice Hulse, Nandita Dukkipati, Adam Paszke, Andrew Bolt, Elnaz Davoodi, Kiam Choo, Jennifer Beattie, Jennifer Prendki, Harsha Vashisht, Rebeca Santamaria-Fernandez, Luis~C. Cobo, Jarek Wilkiewicz, David Madras, Ali Elqursh, Grant Uy, Kevin Ramirez, Matt Harvey, Tyler Liechty, Heiga Zen, Jeff Seibert, Clara~Huiyi Hu, Mohamed Elhawaty, Andrey Khorlin, Maigo Le, Asaf Aharoni, Megan Li, Lily Wang, Sandeep Kumar, Alejandro Lince, Norman Casagrande, Jay Hoover, Dalia~El Badawy, David Soergel, Denis Vnukov, Matt Miecnikowski, Jiri Simsa, Anna Koop, Praveen Kumar, Thibault Sellam, Daniel Vlasic, Samira Daruki, Nir Shabat, John Zhang, Guolong Su, Jiageng Zhang, Jeremiah Liu, Yi~Sun, Evan Palmer, Alireza Ghaffarkhah, Xi~Xiong, Victor Cotruta, Michael Fink, Lucas Dixon,
  Ashwin Sreevatsa, Adrian Goedeckemeyer, Alek Dimitriev, Mohsen Jafari, Remi Crocker, Nicholas FitzGerald, Aviral Kumar, Sanjay Ghemawat, Ivan Philips, Frederick Liu, Yannie Liang, Rachel Sterneck, Alena Repina, Marcus Wu, Laura Knight, Marin Georgiev, Hyo Lee, Harry Askham, Abhishek Chakladar, Annie Louis, Carl Crous, Hardie Cate, Dessie Petrova, Michael Quinn, Denese Owusu-Afriyie, Achintya Singhal, Nan Wei, Solomon Kim, Damien Vincent, Milad Nasr, Christopher~A. Choquette-Choo, Reiko Tojo, Shawn Lu, Diego de~Las~Casas, Yuchung Cheng, Tolga Bolukbasi, Katherine Lee, Saaber Fatehi, Rajagopal Ananthanarayanan, Miteyan Patel, Charbel Kaed, Jing Li, Jakub Sygnowski, Shreyas~Rammohan Belle, Zhe Chen, Jaclyn Konzelmann, Siim Põder, Roopal Garg, Vinod Koverkathu, Adam Brown, Chris Dyer, Rosanne Liu, Azade Nova, Jun Xu, Slav Petrov, Demis Hassabis, Koray Kavukcuoglu, Jeffrey Dean, and Oriol Vinyals.
\newblock Gemini 1.5: Unlocking multimodal understanding across millions of tokens of context, 2024.

\bibitem[Sheng et~al.(2023)Sheng, Zheng, Yuan, Li, Ryabinin, Chen, Liang, R{\'e}, Stoica, and Zhang]{sheng2023flexgen}
Ying Sheng, Lianmin Zheng, Binhang Yuan, Zhuohan Li, Max Ryabinin, Beidi Chen, Percy Liang, Christopher R{\'e}, Ion Stoica, and Ce~Zhang.
\newblock Flexgen: High-throughput generative inference of large language models with a single gpu.
\newblock In \emph{International Conference on Machine Learning}, pp.\  31094--31116. PMLR, 2023.

\bibitem[Taori et~al.(2023)Taori, Gulrajani, Zhang, Dubois, Li, Guestrin, Liang, and Hashimoto]{alpaca}
Rohan Taori, Ishaan Gulrajani, Tianyi Zhang, Yann Dubois, Xuechen Li, Carlos Guestrin, Percy Liang, and Tatsunori~B. Hashimoto.
\newblock Stanford alpaca: An instruction-following llama model.
\newblock \url{https://github.com/tatsu-lab/stanford_alpaca}, 2023.

\bibitem[Touvron et~al.(2023)Touvron, Martin, Stone, Albert, Almahairi, Babaei, Bashlykov, Batra, Bhargava, Bhosale, et~al.]{touvron2023llama}
Hugo Touvron, Louis Martin, Kevin Stone, Peter Albert, Amjad Almahairi, Yasmine Babaei, Nikolay Bashlykov, Soumya Batra, Prajjwal Bhargava, Shruti Bhosale, et~al.
\newblock Llama 2: Open foundation and fine-tuned chat models.
\newblock \emph{arXiv preprint arXiv:2307.09288}, 2023.

\bibitem[Vaswani et~al.(2017)Vaswani, Shazeer, Parmar, Uszkoreit, Jones, Gomez, Kaiser, and Polosukhin]{vaswani2017attention}
Ashish Vaswani, Noam Shazeer, Niki Parmar, Jakob Uszkoreit, Llion Jones, Aidan~N Gomez, {\L}ukasz Kaiser, and Illia Polosukhin.
\newblock Attention is all you need.
\newblock \emph{Advances in neural information processing systems}, 30, 2017.

\bibitem[Wolf et~al.(2020)Wolf, Debut, Sanh, Chaumond, Delangue, Moi, Cistac, Rault, Louf, Funtowicz, Davison, Shleifer, von Platen, Ma, Jernite, Plu, Xu, Scao, Gugger, Drame, Lhoest, and Rush]{wolf-etal-2020-transformers}
Thomas Wolf, Lysandre Debut, Victor Sanh, Julien Chaumond, Clement Delangue, Anthony Moi, Pierric Cistac, Tim Rault, Rémi Louf, Morgan Funtowicz, Joe Davison, Sam Shleifer, Patrick von Platen, Clara Ma, Yacine Jernite, Julien Plu, Canwen Xu, Teven~Le Scao, Sylvain Gugger, Mariama Drame, Quentin Lhoest, and Alexander~M. Rush.
\newblock Transformers: State-of-the-art natural language processing.
\newblock In \emph{Proceedings of the 2020 Conference on Empirical Methods in Natural Language Processing: System Demonstrations}, pp.\  38--45, Online, October 2020. Association for Computational Linguistics.
\newblock URL \url{https://www.aclweb.org/anthology/2020.emnlp-demos.6}.

\bibitem[Xia et~al.(2023)Xia, Gao, Zeng, and Chen]{xia2023sheared}
Mengzhou Xia, Tianyu Gao, Zhiyuan Zeng, and Danqi Chen.
\newblock Sheared llama: Accelerating language model pre-training via structured pruning.
\newblock \emph{arXiv preprint arXiv:2310.06694}, 2023.

\bibitem[Xiao et~al.(2023)Xiao, Lin, Seznec, Wu, Demouth, and Han]{xiao2023smoothquant}
Guangxuan Xiao, Ji~Lin, Mickael Seznec, Hao Wu, Julien Demouth, and Song Han.
\newblock Smoothquant: Accurate and efficient post-training quantization for large language models.
\newblock In \emph{International Conference on Machine Learning}, pp.\  38087--38099. PMLR, 2023.

\bibitem[Yu et~al.(2022)Yu, Jeong, Kim, Kim, and Chun]{yu2022orca}
Gyeong-In Yu, Joo~Seong Jeong, Geon-Woo Kim, Soojeong Kim, and Byung-Gon Chun.
\newblock Orca: A distributed serving system for $\{$Transformer-Based$\}$ generative models.
\newblock In \emph{16th USENIX Symposium on Operating Systems Design and Implementation (OSDI 22)}, pp.\  521--538, 2022.

\bibitem[Yuan et~al.(2024)Yuan, Shang, Zhou, Dong, Zhou, Xue, Wu, Li, Gu, Lee, Yan, Chen, Sun, and Keutzer]{yuan2024llm}
Zhihang Yuan, Yuzhang Shang, Yang Zhou, Zhen Dong, Zhe Zhou, Chenhao Xue, Bingzhe Wu, Zhikai Li, Qingyi Gu, Yong~Jae Lee, Yan Yan, Beidi Chen, Guangyu Sun, and Kurt Keutzer.
\newblock Llm inference unveiled: Survey and roofline model insights, 2024.

\bibitem[Zhong et~al.(2024)Zhong, Liu, Chen, Hu, Zhu, Liu, Jin, and Zhang]{zhong2024distserve}
Yinmin Zhong, Shengyu Liu, Junda Chen, Jianbo Hu, Yibo Zhu, Xuanzhe Liu, Xin Jin, and Hao Zhang.
\newblock Distserve: Disaggregating prefill and decoding for goodput-optimized large language model serving.
\newblock \emph{arXiv preprint arXiv:2401.09670}, 2024.

\end{thebibliography}
\bibliographystyle{colm2024_conference}
\newpage
\section{Appendix}

\section{Mean GPU utilization comparison}
\begin{figure}[h]
\begin{center}
    \includegraphics[width=\textwidth]{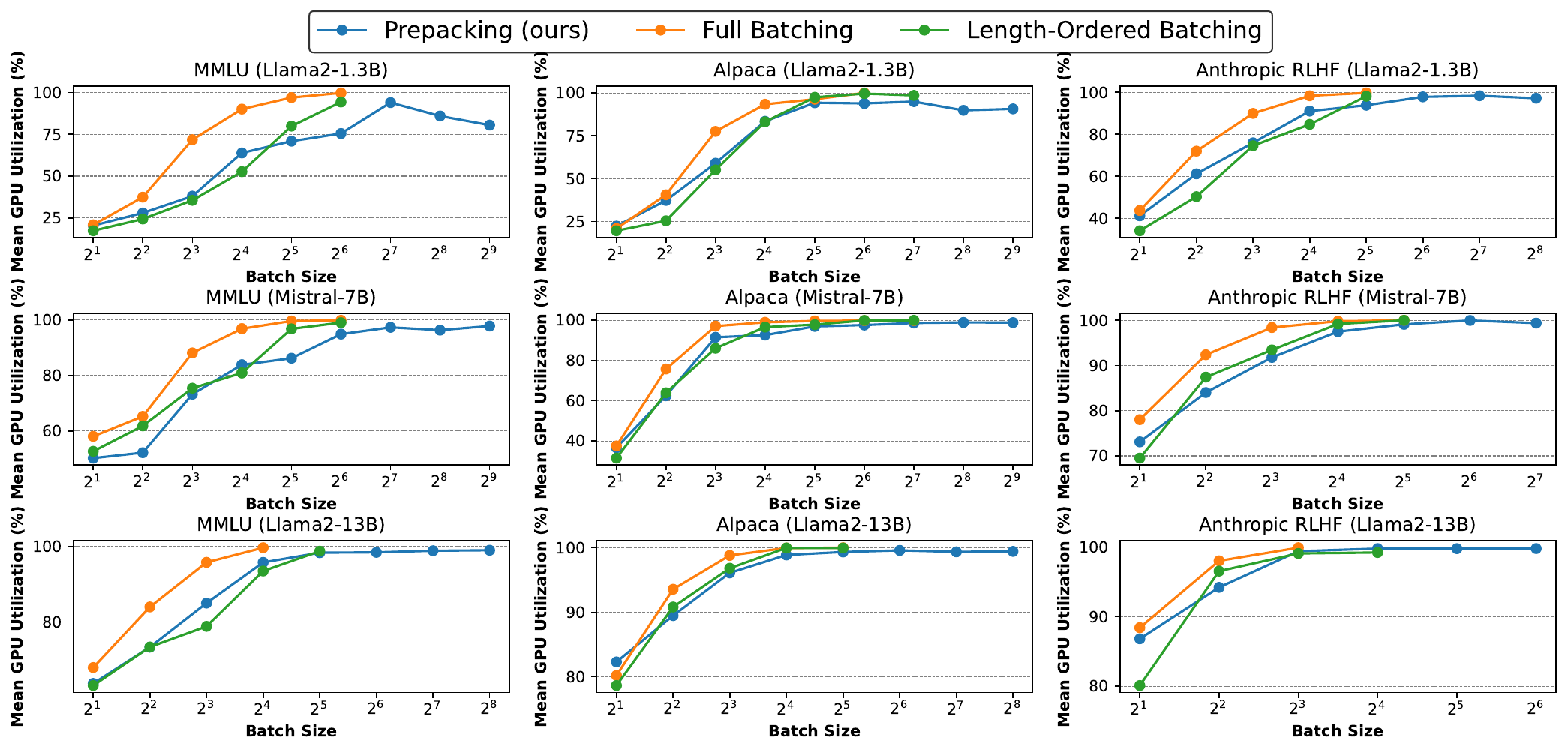}
    \caption{Mean GPU utilization for prefilling the prompts in datasets, sampled with a fixed batch size. \cname{} achieves lightest GPU utilization when the batch size is the same for every method.}
    \label{meangpuutil}
    \end{center}
\end{figure}

\section{Dataset length distribution statistics}
\begin{table}[ht]
\centering
\caption{Evaluation Datasets Length Statistics. Due to computational resources constraints, we choose subsets from these datasets for evaluation.}
\scalebox{0.85}{
\begin{tabular}{lccc}
\hline
\textbf{Dataset name} & \textbf{Subset Min. / Mean / Max. SeqLength} \\
\hline
Wikitext Max SeqLen 256 (Wiki256)~\citep{merity2016pointer} & 1 / 73 / 256 \\
Wikitext Max SeqLen 512 (Wiki512)~\citep{merity2016pointer} & 6 / 120 / 512 \\
MMLU~\citep{hendryckstest2021} & 4 / 64 / 1102  \\
Anthropic HH RLHF~\citep{bai2022training} & 22 / 247 / 1620  \\
Alpaca~\citep{alpaca} & 43 / 126 / 527  \\
SamSum~\citep{gliwa-etal-2019-samsum} & 21 / 169 / 942  \\

\hline
\end{tabular}
\label{tab:datasets_length}
}
\end{table}

\section{Model details}
\begin{table}[ht]
\centering
\caption{Model architecture used in the evaluations}
\scalebox{0.86}{
\begin{tabular}{lcccc}
\hline
\textbf{Model} & \textbf{Num Params} & \textbf{Num layers} & \textbf{Hidden dim} & \textbf{Num heads} \\
\hline
Sheared LLAMA 1.3B~\citep{xia2023sheared} & 1.3B & 24 & 2048 & 16 \\
LLAMA 2 7B~\citep{touvron2023llama}  & 7B & 32 & 4096 & 32 \\
Mistral 7B~\citep{jiang2023mistral} & 7B & 32 & 4096 & 32 \\
LLAMA 2 13B~\citep{touvron2023llama}  & 13B & 40 & 4096 & 40 \\

\hline
\end{tabular}
\label{tab:model_architecture}
}
\end{table}

\section{How does the performance gain scale with characteristics of lengths within a batch?}
We extend our runtime analysis from section~\ref{sec:regerssion}, evaluating \name{}'s speedup across various settings using a synthetic dataset with prompt lengths uniformly distributed. The experiments, conducted with the Llama2 1.3B and Llama2 7B, aim to quantify the efficiency gains through Batch Size Reduction ($r/k$) and Max Absolute Deviation ($m - L/k$). These findings, presented in detailed plots, offer insights into \name{}'s performance scalability.
\label{sec:statisticsplot_more}
\begin{figure}[h]
\begin{center}
    \includegraphics[width=\textwidth]{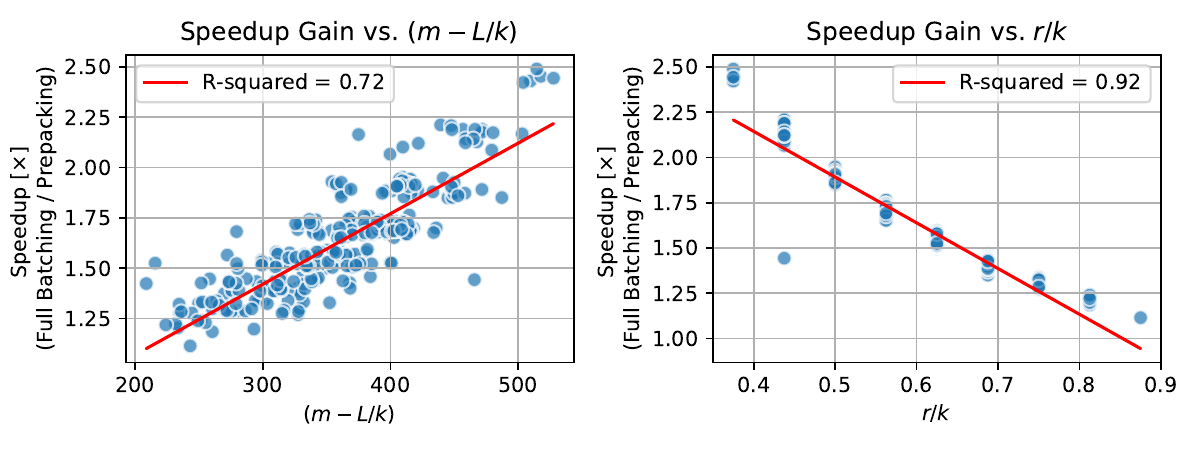}
    \caption{Speedup gains relative to full batching, with respect to Batch Size Reduction ($r/k$) and Max Absolute Deviation ($m - L/k$), conducted on Llama2 1.3B with batch size 16 and 5000 prompts.}
    \label{meangpuutil}
    \end{center}
\end{figure}
\begin{figure}[h]
\begin{center}
    \includegraphics[width=\textwidth]{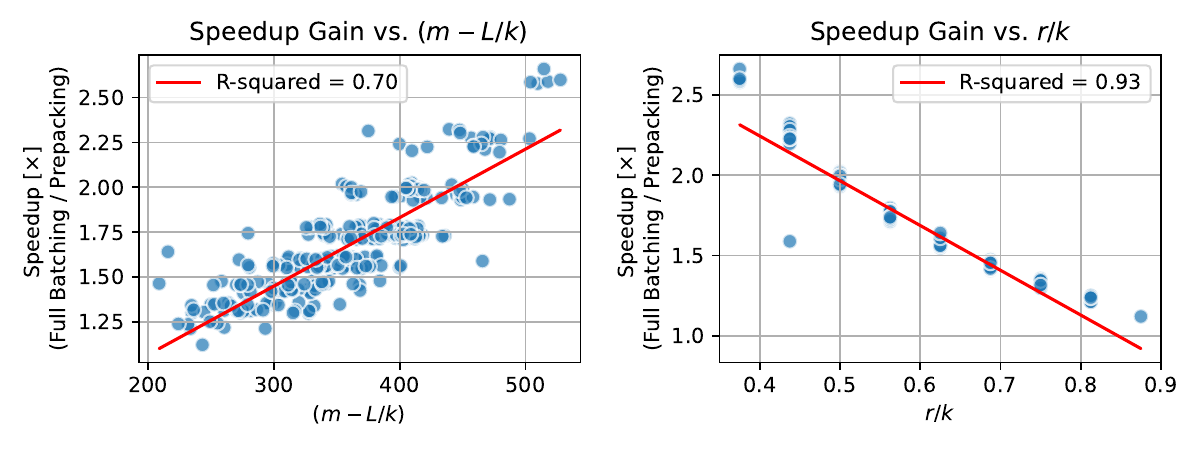}
    \caption{Speedup gains relative to full batching, with respect to Batch Size Reduction ($r/k$) and Max Absolute Deviation ($m - L/k$), conducted on Llama2 7B with batch size 16 and 2500 prompts.}
    \label{meangpuutil}
    \end{center}
\end{figure}
\begin{figure}[h]
\begin{center}
    \includegraphics[width=\textwidth]{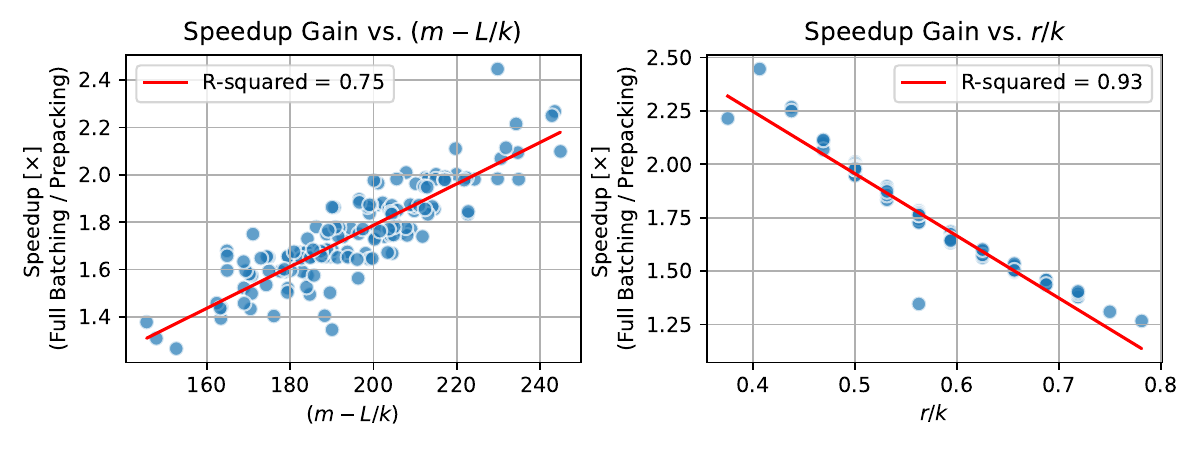}
    \caption{Speedup gains relative to full batching, with respect to Batch Size Reduction ($r/k$) and Max Absolute Deviation ($m - L/k$), conducted on Llama2 7B with batch size 32 and 5000 prompts.}
    \label{meangpuutil}
    \end{center}
\end{figure}
\begin{figure}[h]
\begin{center}
    \includegraphics[width=\textwidth]{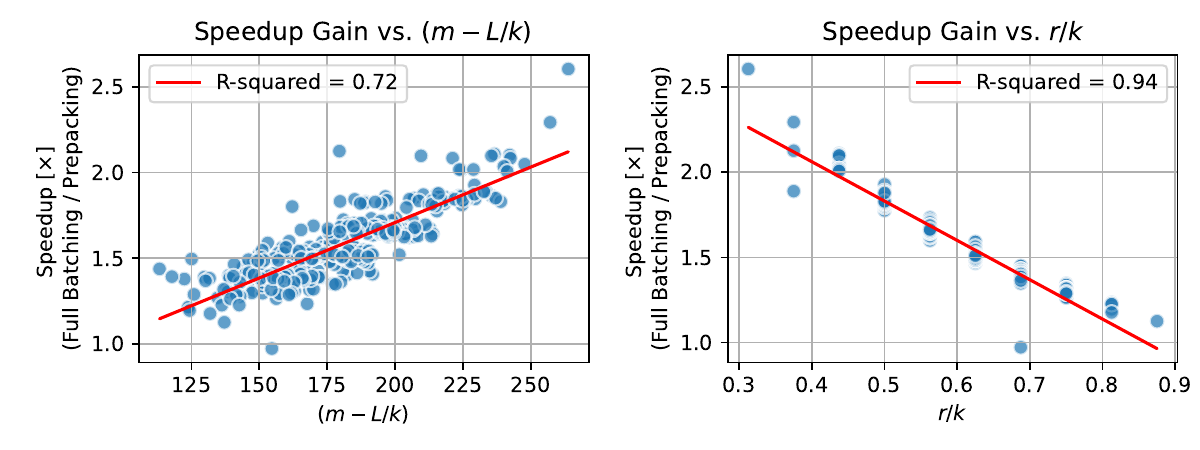}
    \caption{Speedup gains relative to full batching, with respect to Batch Size Reduction ($r/k$) and Max Absolute Deviation ($m - L/k$), conducted on Mistral 7B with batch size 16 and 5000 prompts.}
    \label{meangpuutil}
    \end{center}
\end{figure}

\end{document}